\documentclass[11pt]{article} 
\usepackage{setspace}
\usepackage[margin=1in]{geometry}

\usepackage[utf8]{inputenc} 
\usepackage[T1]{fontenc}    
\usepackage{footnote}
\usepackage{hyperref}       
\usepackage{url}            
\usepackage{booktabs}       
\usepackage{amsfonts}       
\usepackage{nicefrac}       
\usepackage{microtype}      
\usepackage{xcolor,graphicx}         
\usepackage{subcaption}
\usepackage{multirow}
\usepackage{caption}
\usepackage{comment}
\usepackage{float}
\usepackage{pdflscape}
\usepackage{enumitem}
\usepackage{natbib}
\usepackage{amsmath}
\usepackage[capitalise, noabbrev]{cleveref}
\usepackage{rotating}
\usepackage[normalem]{ulem}
\useunder{\uline}{\ul}{}

\usepackage{array}       
\usepackage{caption}     

\renewcommand{\arraystretch}{1.05}
\newcommand{\Var}{\mathbb{V}}

\newcommand{\Cov}{\mathrm{Cov}}

\usepackage{booktabs}
\title{Demystifying Prediction Powered Inference}
\usepackage{authblk}

\author[1]{Yilin Song}
\author[2]{Dan M. Kluger}
\author[3]{Harsh Parikh\thanks{Co-corresponding author. Email: harsh.parikh@yale.edu}}
\author[1]{Tian Gu\thanks{Co-corresponding author. Email: tg2880@cumc.columbia.edu}}

\affil[1]{Department of Biostatistics, 
  Columbia Mailman School of Public Health}
\affil[2]{Institute for Data, Systems, and Society, Massachusetts Institute of Technology}
\affil[3]{Department of Biostatistics, 
  Yale School of Public Health}

\usepackage{pifont}

\definecolor{efficiencyblue}{HTML}{0072B2}
\definecolor{aoneorange}{HTML}{D55E00}
\definecolor{atwogreen}{HTML}{009E73}
\definecolor{athreepurple}{HTML}{CC79A7}

\newcommand{\AxisCheck}[1]{{\large\textcolor{#1}{\ding{51}}}}

\date{}   

\begin{document}

\maketitle
\begin{abstract}
Machine learning predictions are increasingly used to supplement incomplete or costly-to-measure outcomes in fields such as biomedical research, environmental science, and social science. However, treating predictions as ground truth introduces bias while ignoring them wastes valuable information. Prediction-Powered Inference (PPI) offers a principled framework that leverages predictions from large unlabeled datasets to improve statistical efficiency while maintaining valid inference through explicit bias correction using a smaller labeled subset. Despite its potential, the growing PPI variants and the subtle distinctions between them have made it challenging for practitioners to determine when and how to apply these methods responsibly. This paper demystifies PPI by synthesizing its theoretical foundations, methodological extensions, connections to existing statistics literature, and diagnostic tools into a unified practical workflow. 
Using the MOSAIKS housing price data, we show that PPI variants produce tighter confidence intervals than complete-case analysis, but that double-dipping, i.e. reusing training data for inference, leads to anti-conservative confidence intervals and below-nominal coverage. Under missing-not-at-random mechanisms, all methods, including classical inference using only labeled data, yield biased estimates. We provide a decision flowchart linking assumption violations to appropriate PPI variants, a summary table of representative methods, and practical diagnostic strategies for evaluating core assumptions. By framing PPI as a general recipe rather than a single estimator, this work bridges methodological innovation and applied practice, helping researchers responsibly integrate predictions into valid inference.
\end{abstract}


\section{Introduction}

\subsection{Why Prediction-Powered Inference?}

Scientific research has persistently faced the challenge that gold-standard measurements are expensive, invasive, or logistically burdensome to collect at scale, while the statistical power required to detect meaningful effects often demands large sample sizes. This tension pervades biomedical research, where imaging phenotypes or validated clinical endpoints may be available for only a fraction of study participants \citep{lou2021leveraging}; environmental science, where ground-truth measurements require costly field campaigns \citep{rolf2021a}; and social science, where high-quality survey responses are increasingly difficult to obtain \citep{meyer2015household, williams2018trends}. This has resulted in a growing body of studies with partially observed outcomes---large datasets in which the quantity of interest is measured for some units but is missing for many others.

Simultaneously, advances in machine learning (ML) have produced increasingly accurate predictive models capable of imputing unmeasured outcomes from readily available covariates \citep{rolf2021a, mccaw2024synthetic, chen2025unified}. Pre-trained models leveraging biobanks, electronic health records \citep{gu2023commute,mccaw2024synthetic}, satellite imagery \citep{rolf2021a, proctor2023parameter}, or large language models (LLMs) \citep{StratifiedPPI} can generate predictions for nearly any unit with observed features. These predictions often carry substantial information about the true outcomes, raising a natural question: {how can investigators leverage ML predictions to improve statistical inference without sacrificing validity?}

However, treating predictions as ground truth risks systematic bias whenever the predictive model is imperfect. Confidence intervals may be narrow but centered on the wrong quantity and suffer from poor inferential guarantees. In contrast, ignoring predictions entirely and analyzing only labeled cases maintains validity but discards potentially valuable information, resulting in wider confidence intervals and reduced power. Neither extreme is satisfactory.

Prediction-Powered Inference (PPI), introduced by \citet{PPI}, offers a principled middle path. The key insight is elegantly simple: use predictions from a large unlabeled sample to boost efficiency, but explicitly correct for prediction error using a smaller labeled subset where both predictions and true outcomes are observed. This bias-correction mechanism ensures that inferences remain valid regardless of prediction quality, with poor predictions merely failing to improve efficiency rather than compromising validity. When predictions are informative, PPI can yield substantially tighter confidence intervals than complete-case (CC) analysis while maintaining nominal coverage.

{As a concrete example, consider a housing dataset where satellite imagery is used to predict housing prices across a large number of locations while observed housing prices are available for only a subset of properties. PPI combines these predictions with the observed prices to estimate quantities such as the association between housing price and neighborhood characteristics, improving statistical efficiency without sacrificing validity. However, this validity depends on using predictions generated independently of the inference data. If the same housing observations are used both to train the prediction model and to estimate the regression coefficients with PPI, the prediction errors become artificially optimistic, leading to confidence intervals that are too narrow. This phenomenon, known as double-dipping, is one of the most important practical pitfalls in applying PPI. In Section \ref{sec:DataBasedExperiments}, we use the MOSAIKS housing price dataset to demonstrate both the efficiency gains achievable with PPI and the consequences of violating this assumption.}

{\subsection{Representative Applications of Prediction-Powered Inference}

Although PPI is built upon a unified statistical framework, its applications span a wide range of scientific disciplines and inferential goals. Across these settings, the common challenge is that gold-standard outcomes are available for only a small subset of observations, while machine learning models can generate predictions for a much larger population. PPI combines these two sources of information to improve statistical efficiency while preserving valid statistical inference. Below, we present several representative scenarios illustrating when PPI is most naturally applicable. These examples are intended to motivate the framework rather than provide an exhaustive review of the rapidly growing PPI literature.

\underline{Biomedical studies: estimating disease prevalence and risk factors.}
Large cohort studies often aim to estimate disease prevalence, biomarker distributions, or associations between patient characteristics and disease outcomes. Gold-standard outcomes may require expert adjudication, medical imaging, laboratory assays, or manual chart review, making complete labeling prohibitively expensive. Meanwhile, prediction models based on electronic health records or medical imaging can generate outcome predictions for the entire cohort. The inferential targets are typically population means (e.g., disease prevalence) or regression coefficients. PPI combines predictions from the full cohort with gold-standard labels from a smaller validation subset to improve statistical efficiency while maintaining valid inference.

\underline{Environmental science: large-scale monitoring from remote sensing.}
Environmental studies often seek to estimate regional environmental quantities or quantify relationships between environmental outcomes and geographic or socioeconomic factors. Ground-truth measurements require costly field surveys, whereas satellite imagery and remote sensing models provide predictions at nearly every location. The inferential targets commonly include population means, regression coefficients, and spatial or temporal trends. PPI combines abundant predictions with a limited number of field measurements to improve statistical efficiency while correcting for prediction bias, substantially reducing the need for expensive data collection. 


\underline{Human annotation and large language model evaluation.}
Evaluating large language models and analyzing large text corpora often require expensive human annotations, such as ratings of factuality, helpfulness, or toxicity. Modern language models can provide inexpensive surrogate evaluations at scale, but these predictions cannot be treated as ground truth. The inferential targets typically include average human evaluation scores, differences between competing models, or regression coefficients relating text characteristics to human judgments. PPI combines a small number of human annotations with abundant machine-generated evaluations to obtain valid and more efficient inference while substantially reducing annotation costs. Similar ideas also arise in large-scale social science studies that rely on expert annotation of text or other unstructured data \citep{emmenegger2026manytasks}.

These examples illustrate that PPI is broadly applicable whenever a relatively small labeled sample containing both gold-standard outcomes and predictions is accompanied by a substantially larger sample with predictions alone. Rather than being tied to a particular scientific domain or prediction model, PPI is defined by this data structure and inferential objective. Across these settings, PPI provides a principled framework for obtaining statistically valid and more efficient inference for a broad class of target parameters, including population means, regression coefficients, treatment effects, and other M-estimation problems.}

\subsection{Challenges in Applying PPI}

{Despite its conceptual elegance and practical promise, the growing number of PPI variants and their subtle methodological differences can make it difficult for practitioners outside the statistics and machine learning communities to determine when and how to apply these methods appropriately.} We identify three barriers impeding broader adoption:

\begin{enumerate}[leftmargin=*]
    \item \textbf{Conceptual barriers:} When is PPI appropriate for a given study? What assumptions underpin its validity, and how restrictive are they in practice? What are the connections with existing literature? The answers are scattered across technical papers that assume substantial background in statistical 
    theory.
    
    \item \textbf{Operational barriers:} How should practitioners implement PPI? What data structures are required? Which variant among the growing family of PPI methods should one choose? How can assumptions be assessed, or at least interrogated, with available data?
    
    \item \textbf{Interpretational barriers:} How should PPI results be communicated? What do efficiency gains (or their absence) imply about the underlying prediction model? What are the consequences of assumption violations, and how can practitioners avoid common pitfalls such as ``double-dipping'', using overlapping data for model training and inference?
\end{enumerate}

\subsection{Contributions and Outline}
This paper addresses these gaps by demystifying PPI and presenting a unified, accessible framework that synthesizes its theoretical foundations, methodological variants, and practical diagnostics. Our exposition is organized around resolving each of the three barriers:

\begin{enumerate}[leftmargin=*]
    \item 
    We articulate the core identification assumptions---distributional comparability between labeled and unlabeled samples, independence of the pre-trained model from inference data, and complete covariate availability---and describe how different PPI variants relax or modify these requirements (Section~\ref{sec:ppi-framework}). In Section \ref{sec:ppi_connection}, we discuss its theoretical foundations and connections with existing statistics literature.
    
    \item 
    We provide a step-by-step workflow for implementation, including data preparation, assumption diagnostics, method selection, and result validation (Section~\ref{sec:guideline}).
    
    \item 
    {We offer practical guidance on interpreting and reporting PPI analyses, including how to communicate efficiency gains, document assumptions, compare PPI with complete-case analyses, and recognize situations in which apparent precision may be misleading (Section~\ref{sec:guideline} and \ref{sec:discussion}).}
\end{enumerate}

In Section~\ref{sec:DataBasedExperiments}, using the MOSAIKS housing dataset \citep{rolf2021a}, we empirically reveal both the promise and the pitfalls of PPI. 
{The MOSAIKS dataset is well suited for illustrating PPI methods because it contains both observed outcomes and corresponding machine-learning predictions, allowing direct comparison among PPI estimators, classical estimators, and oracle procedures. This enables us to systematically evaluate the impact of different assumptions and implementation choices while retaining access to ground-truth quantities. In addition, the predictive performance in this dataset lies in a practically informative regime: the predictions are sufficiently accurate to yield meaningful efficiency gains, yet imperfect enough that differences among PPI methods remain visible. When predictions are nearly perfect, most PPI variants behave similarly and have consistent gains over the CC analysis; when predictions are highly inaccurate, the predictions contribute little information and the advantages of PPI diminish. The MOSAIKS dataset therefore provides a realistic and transparent benchmark for demonstrating the strengths, limitations, and practical behavior of different PPI approaches.}
In this application, we also show that performance deteriorates under violations such as “double-dipping” and missing-not-at-random (MNAR) mechanisms, emphasizing the importance of satisfying PPI's core assumptions.

Beyond the methodological synthesis, we contribute several practical resources: a decision flowchart linking assumption violations to appropriate PPI variants (Figure~\ref{fig:recommendation}), diagnostic strategies for evaluating assumptions in applied settings (Table~\ref{tab:assumption_diagnostics}), and a summary table of representative methods with their core assumptions and recommended use cases (Table~\ref{tab:ppi_summary}).

By framing PPI as a general recipe rather than a single estimator, this paper aims to bridge methodological innovation and applied practice. We hope to enable researchers across disciplines to responsibly leverage the predictive power of modern ML for more efficient and valid statistical inference.

\begin{figure}[htp]
    \centering
    \includegraphics[width=.9\linewidth]{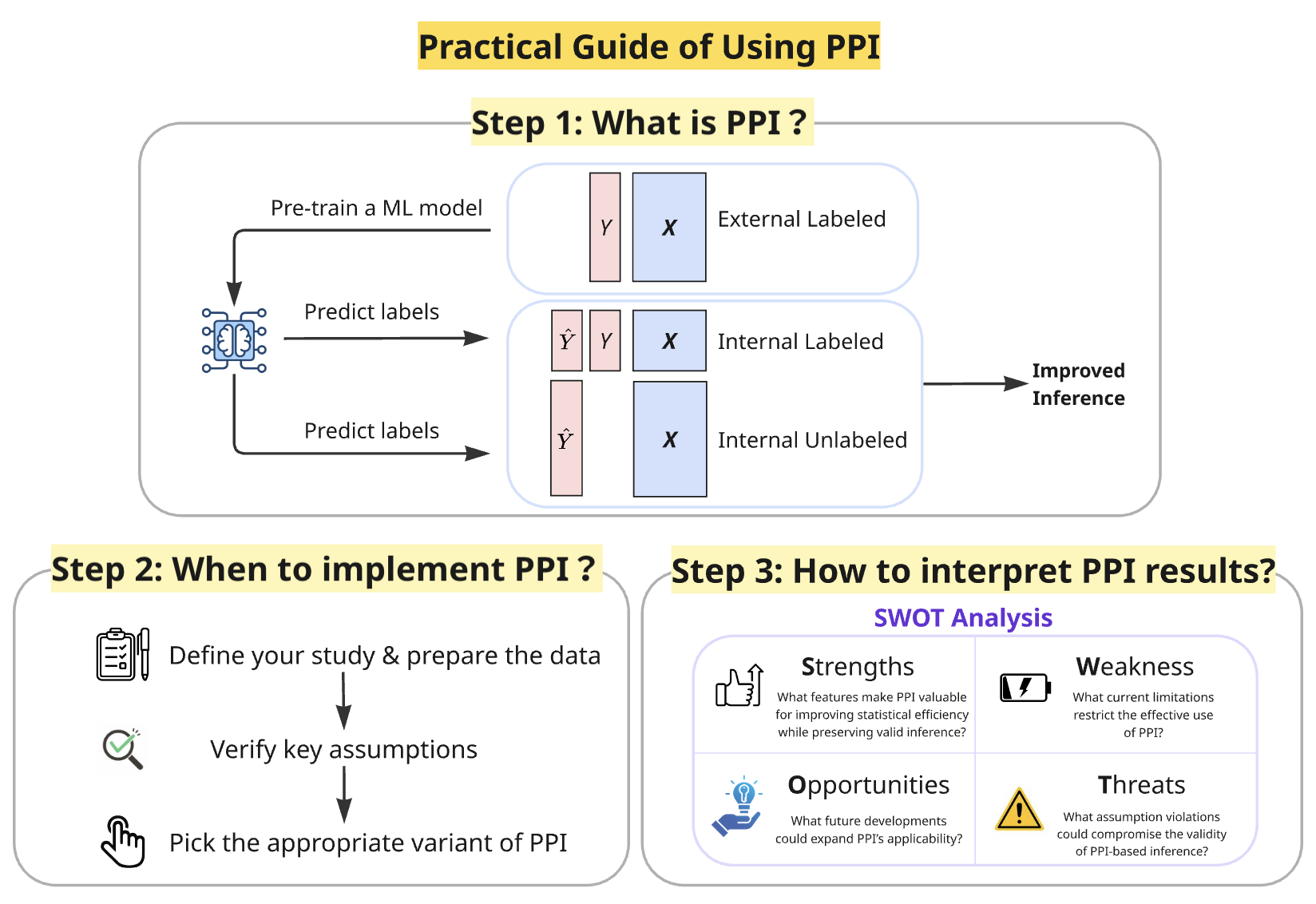}
    \caption{Overall workflow of the practical guide of using PPI.}
    \label{fig:workflow}
\end{figure}

\section{Prediction-Powered Inference: Framework and Extensions}
\label{sec:ppi-framework}

PPI is designed for studies where the outcome of interest is difficult or costly to measure for every unit, but prediction models are available for a much larger unlabeled sample. A canonical example arises in clinical studies: baseline covariates $X_i$ (e.g., demographics, medical history, or lab results) are observed for all $i=1, \dots, n$ participants, while the gold-standard outcome $Y_i$ (e.g., frailty or imaging phenotype) is ascertained only for a subset due to budget or logistical constraints. At the same time, an external or separately trained model can produce predictions $\hat Y_i$ given $X_i$ for all units. 

Traditional complete case analysis uses only those individuals with observed $Y_i$, discarding unlabeled units and any predictive information in $\hat Y_i$. In contrast, PPI treats predictions as an auxiliary resource: it uses them to boost efficiency while explicitly correcting for prediction error, so that inferences still target the parameter defined in terms of the true outcome.

\subsection{Basic setup and estimator}

Assume a pre-trained prediction model $\hat f$ is obtained from an external dataset. In the internal data, some units have observed outcomes while others do not. Let $S_i$ denote an indicator of outcome availability, with $S_i=1$ if $Y_i$ is observed (labeled) and $S_i=0$ otherwise (unlabeled). For all $n$ subjects, we observe $(X_i, S_i)$, and we observe $Y_i$ only when $S_i=1$. Predicted outcomes $\hat Y_i=\hat f(X_i)$ are computed for all units using the external model (Figure~\ref{fig:workflow}, Step~1).

{Let $\theta^*$ be the target parameter, defined as the minimizer of a population loss,
\[
\theta^\star = \arg\min_{\theta \in \Theta} \, \mathbb{E}\{\ell(Y,X;\theta)\},
\]
where $\ell$ is a user-chosen loss function (e.g., squared error for mean estimation,
negative log-likelihood for generalized linear models, or a risk function for
classification) and $\Theta$ is a parameter space (often $\Theta=\mathbb{R}^d$ for some positive integer $d$)\footnote{Standard M-estimation conditions are imposed: $\ell(\cdot,\cdot\,;\theta)$
is convex and (almost everywhere) differentiable in $\theta$, and $\theta^\star$ is the unique minimizer, lying
in the interior of $\Theta$, with a positive-definite Hessian
$H_{\theta^\star} = \nabla_\theta^2\, \mathbb{E}\{\ell(Y,X;\theta)\} \big|_{\theta=\theta^*}$, while both $\theta \mapsto \ell(Y,X;\theta)$ and $\theta \mapsto \ell(\hat{Y},X;\theta)$ are locally Lipschitz in a neighborhood of $\theta^*$; see \citet{PPI++} for more details.}.  
}

A CC estimator based only on internal labeled units solves
\[
\hat\theta_{\mathrm{CC}}
= \arg\min_{\theta \in \Theta} \frac{1}{n_\ell}\sum_{i : S_i = 1} \ell(Y_i, X_i; \theta),
\]
where $n_\ell = \sum_i S_i$ is the labeled sample size. The PPI estimator augments this objective with predictions from internal unlabeled units and then subtracts a bias correction estimated on all internal labeled data. A convenient generic form is
\begin{equation}
\label{eq:ppi-estimator}
\hat\theta_{\mathrm{PPI}}
= \arg\min_{\theta \in \Theta}
\left[
\frac{1}{n_u}\sum_{i : S_i = 0} \ell(\hat Y_i, X_i; \theta)
\;-\;
\frac{1}{n_\ell}\sum_{i : S_i = 1} \big\{\ell(\hat Y_i, X_i; \theta) - \ell(Y_i, X_i; \theta)\big\}
\right],
\end{equation}
where $n_u = n - n_\ell$ is the unlabeled sample size. The first term uses predictions in place of missing outcomes to approximately evaluate the average loss for unlabeled units; the second term estimates and subtracts the average error in the prediction-based estimates of the loss using internal labeled units, thereby giving a debiased empirical loss. This structure mirrors the augmented bias correction estimator in the missing completely at random (MCAR) setting: a plug-in term based on predicted outcomes is corrected by an influence-function style augmentation term computed on the internal labeled units, yielding an estimator that combines outcome modeling with a bias-removal adjustment \citep{Robins1994}. %

In the special case of mean estimation with squared loss, i.e., $\ell(Y,X;\theta)=(Y-\theta)^2$, Equation \eqref{eq:ppi-estimator} reduces to a simple decomposition:
\[
\hat\theta_{\mathrm{PPI}}
= \underbrace{\frac{1}{n_u}\sum_{i:S_i=0} \hat Y_i}_{\text{prediction mean}}
\;+\;
\underbrace{\frac{1}{n_\ell}\sum_{i : S_i = 1} (Y_i - \hat Y_i)}_{\text{residual mean}},
\]
the overall mean of predictions on the unlabeled data plus the average residual on labeled data. 

{Alternatively}, in many variants of PPI (e.g., \cite{ZrnicActiveInference,gronsbell2024another}) the first term in Equation \eqref{eq:ppi-estimator} involves summing over all $i \in \{1,\dots,n\}$ as opposed to only elements where $S_i=0$, resulting instead in the following mean estimator \[
\hat\theta_{\mathrm{PPI}}
= \frac{1}{n}\sum_{i=1}^n \hat Y_i
\;+\;
\frac{1}{n_\ell}\sum_{i : S_i = 1} (Y_i - \hat Y_i).
\] 
{When the labels $Y$ are MCAR, the choice of whether the first term sums over all samples or just those with $S_i=0$ has no impact on the bias (both first terms' means are the same) and has a negligible impact on the variance of the estimator (especially when $n_\ell \ll n$). Moreover, under efficiency-guaranteeing refinements (see Section \ref{sec:efficiencyRefinements}) the two approaches result in the same asymptotic variance (e.g., see Proposition F.3 of \cite{Kluger25GeneralizingPPI}). When the labels $Y$ are instead missing at random (MAR) but are not MCAR, the variant in which the first term averages over only the $n_u$ unlabeled samples rather than all $n$ samples is a bit more complex and less efficient. In particular, the all-sample approach requires fewer terms with inverse-probability-weighted averages, and under efficiency-guaranteeing refinements (see Section \ref{sec:efficiencyRefinements}), the latter has been found to result in smaller asymptotic variance (e.g., see Section B.1.2 of \cite{zou2026generalizedpredictionpoweredinferenceapplication}).} 

\subsection{Core assumptions and efficiency intuition}

PPI is designed to deliver valid inference even when the prediction model is misspecified. Its validity is governed by the following set of assumptions: 
    
\begin{description}
    \item[(A1) Distribution comparability between the labeled and unlabeled set]  
    The internal labeled and unlabeled samples must be drawn from the same population. 
    Thus, the outcomes are MCAR.\footnote{An extension to the known covariate shift or label shift setting was considered in the original PPI work \citep{PPI}.}
    Intuitively, this ensures that labeled residuals are representative; otherwise, the bias correction term cannot debias predictions for the target population.
    \item[(A2) Pre-trained prediction model on external data]  
    There exists a prediction model, $\hat f$, trained on an external labeled dataset that does not overlap with the internal data, which maps covariates $X_i$ to predicted outcomes $\hat Y_i = \hat f(X_i)$. This prevents overfitting to contaminate the bias correction and ensures valid standard error calculations.

    \item[(A3) Complete covariate information]
    All covariates $X$ required by the prediction model and the inferential objective are fully observed for every unit, labeled and unlabeled. This guarantees that the prediction function $\hat Y_i = \hat f(X_i)$ can be evaluated on all unlabeled units.

\end{description}

\subsection{A unified view of PPI variants}

The original PPI estimator in Equation \eqref{eq:ppi-estimator} has inspired a broad family of methods that share a common principle -- {obtain predictions using ML methods and then debias them using labeled outcomes}. However, they differ in how they handle efficiency and the three core assumptions from the original PPI. For practitioners, it is helpful to view these contributions as operating along the following four axes.

\subsubsection{Efficiency-guaranteeing refinements}\label{sec:efficiencyRefinements}

PPI is not automatically more efficient than CC analysis. First, let's characterize  its behavior in the simple mean-estimation setting 
under completely random missingness. Let $\pi = \Pr(S=1)$ denote the labeled fraction. As shown in 
\citet{PPI}, assuming a fixed prediction function $\hat{f}$ and independent 
and identically distributed (IID) observations $\{(X_i, Y_i, S_i)\}_{i=1}^n$, 
a direct calculation yields
\[
\Var(\hat\theta_{\mathrm{PPI}} \mid \hat{f}) - \Var(\hat\theta_{\mathrm{CC}})
\;=\; \frac{\Var(Y - \hat{f}(X)) - \Var(Y)}{\pi n} 
+ \frac{\Var(\hat{f}(X))}{(1-\pi)n}
\;=\; \frac{\Var(\hat{f}(X))}{\pi(1-\pi)n} 
- \frac{2\Cov(Y, \hat{f}(X))}{\pi n}.
\]
When the labeled fraction is small, i.e., $(1-\pi) \approx 1$, this 
simplifies to
\[
\Var(\hat\theta_{\mathrm{PPI}} \mid \hat{f}) - \Var(\hat\theta_{\mathrm{CC}})
\;\approx\; \frac{\Var(\hat{f}(X)) - 2\Cov(Y, \hat{f}(X))}{\pi n}.
\]
Thus, PPI achieves lower variance than CC whenever 
$\Var(\hat{f}(X)) < 2\Cov(Y, \hat{f}(X))$
. 
The two estimators have equal efficiency when 
$\Var(\hat{f}(X)) = 2\Cov(Y, \hat{f}(X))$---for example, when $\hat{f}$ 
is constant---and PPI can be \emph{less} efficient when 
$\Var(\hat{f}(X)) > 2\Cov(Y, \hat{f}(X))$. In practice, informative 
predictions yield genuine efficiency gains, while poor or adversarial 
predictions can negate or even reverse those gains.

Because the original PPI estimator does not guarantee efficiency gains 
over CC, a recent wave of refinements explicitly addresses this limitation. 
PPI++ \citep{PPI++} introduces a tuning parameter 
$\lambda \in [0,1]$ that interpolates between the PPI and CC estimators:
\[
\hat\theta_{\mathrm{PPI++}}(\lambda)
\;=\; \arg \min_{\theta \in \Theta}\,
\Biggl[
\frac{1}{n_u}\sum_{i:\, S_i = 0} \lambda \cdot \ell(\hat{Y}_i, X_i; \theta)
\;-\;
\frac{1}{n_\ell}\sum_{i:\, S_i = 1} 
\Bigl\{\lambda \cdot \ell(\hat{Y}_i, X_i; \theta) 
- \ell(Y_i, X_i; \theta)\Bigr\}
\Biggr],
\]
where $\lambda$ is chosen to minimize asymptotic variance: $\lambda = 1$ 
recovers the original PPI estimator, and $\lambda = 0$ recovers the CC 
estimator. Under mild regularity conditions, PPI++ is asymptotically no 
worse than CC and strictly more efficient whenever the predictions carry 
any signal about $Y$. When predictions are uninformative, the optimal 
$\lambda$ shrinks toward zero, causing PPI++ to behave similarly to CC 
and providing robustness against poor model predictions.

Whereas PPI++ employs a single scalar tuning parameter, subsequent work 
has pursued greater efficiency through richer parameterizations. 
\citet{StratifiedPPI} obtain gains over PPI++ via stratification, 
allowing $\lambda$ to vary across strata. \citet{miaoPSPA} and 
\citet{gan2024prediction} reformulate the optimization problem as a 
system of estimating equations, effectively replacing the scalar 
$\lambda$ with a matrix of tuning parameters. 
Building on the estimator 
debiasing framework of \citet{ChenAndChen2000}, \citet{gronsbell2024another} 
and \citet{Kluger25GeneralizingPPI} similarly employ tuning matrices, 
which we conjecture generally leads to efficiency improvements over PPI++. 
Drawing on ideas from \citet{Robins1994}, \citet{ji2025predictions} 
consider an entire tuning {function}---an infinite-dimensional 
generalization---that replaces every term 
$\lambda \cdot \ell(\hat{Y}_i, X_i; \theta)$ with a flexible function 
$g(\hat{Y}_i, X_i; \theta)$ chosen to minimize asymptotic variance. 
They provide a practical sample-splitting implementation and demonstrate 
empirically that their estimator performs no worse than PPI or PPI++.

\subsubsection{Robustness to labeling mechanism and accounting for distribution shift}

Several extensions reinterpret PPI through the lens of missing data and selection, treating the labeling process as an imposed missing-at-random  mechanism and explicitly requiring a positivity (or overlap) condition. The positivity condition ensures that the probability of being labeled is bounded away from zero across the covariate space, so that each subpopulation defined by $X$ has a non-negligible chance of being observed. For instance, \citet{Kluger25GeneralizingPPI} assumed IID sampling under MAR with known labeling probabilities that satisfy overlap. Similarly, \citet{carlson2025unifyingframeworkrobustefficient} assumed MAR and overlap assumptions while treating labeling probabilities as known, though they argued (without proof) that their semiparametric and doubly robust (DR) procedures could be extended to settings with estimated probabilities. In contrast, \citet{testa2025semisup} relaxed the known-probability requirement by allowing the labeling mechanism to be estimable, still assuming MAR and overlap but allowing overlap to vanish as sample size increases. Their semiparametric framework retains double robustness, ensuring valid inference when either the labeling model or the outcome regression is correctly specified. {Complementary to these robustness-focused extensions, \citet{ying2025covshift} studied prediction-powered inference under MAR with an emphasis on characterizing when predicted outcomes are beneficial.
}
Collectively, these works illustrate promising directions for extending PPI to accommodate distributional shifts, balancing trade-offs between robustness and practical feasibility.

\subsubsection{Beyond pre-trained model}

Assumption (A2) of the original PPI assumes access to a pre-trained predictor independent of the internal labeled sample, an assumption often unrealistic in practice. Cross-PPI addresses this by training predictors within the internal labeled dataset using $K$-fold cross-fitting and relying solely on out-of-fold predictions for inference \citep{CrossPPIPaper_ZrnicCandes}. 

Cross-fitting itself is a long-standing idea in semiparametric statistics, used to obtain valid inference when nuisance components are estimated flexibly \citep{robins1995semiparametric, vdl2003unified}. From this perspective, Cross-PPI naturally parallels semiparametric methods for missing data, 
where sparsely measured outcomes $Y$ are imputed from widely observed covariates $X$ and inference proceeds with cross-fitted nuisance estimates. Cross-PPI reformulated these ideas within the PPI framework, providing an accessible bridge for those more familiar with ML perspectives. We extend the discussion of the connection between PPI and semiparametric theory in the section that follows.

In practice, Cross-PPI removes the need for a pre-trained predictor, though it requires careful fold construction---avoiding data-wide preprocessing or hyperparameter tuning---and its finite-sample efficiency may depend on the choice of $K$. A related approach, the cross-prediction-powered bootstrap (Cross-PPBoot), builds on the same cross-fitting principle but forms confidence intervals through bootstrap rather than the central limit theorem (CLT) \citep{Zrnic2024PPBoot}. It is specifically developed for settings without a pre-trained model and has shown power gains over split-train PPBoot baselines. Recent extensions, such as Tuned-CPPI, further refine the Cross-PPI framework by adjusting the degree to which unlabeled data are used while maintaining the cross-fitting strategy, thereby further reducing reliance on external predictors \citep{sifaou2024semi}. Together, these methods address PPI’s dependence on pre-trained models by learning within the internal labeled data while preserving conditional independence through cross-fitting, with the usual caveats regarding potential leakage and fold selection.

\subsubsection{Handling missingness patterns}

Another practical limitation of PPI is the assumption (A3) that covariates are fully observed for all units; several recent works relax this by combining ML imputation with debiasing so inference remains valid even when covariates (and/or outcomes) are missing. \citet{chen2025unified} developed a unified, pattern-stratified framework under MAR with arbitrary missingness patterns: they ``mask-and-impute'' within complete cases, aggregate across patterns with appropriate weights, and prove asymptotic normality and efficiency dominance over weighted CC analyses---thereby directly addressing blockwise missing covariates without requiring completeness. Complementing this, \citet{Kluger25GeneralizingPPI} analyzes PPI with imputed covariates and outcomes via a ``predict-then-debias'' approach, detailing conditions under which intervals remain valid and highlighting pitfalls (e.g., loss nonconvexity) when naively plugging predicted covariates into PPI. They also provide bootstrap approaches that construct valid confidence intervals without requiring access to existing PPI software or mathematically derived asymptotic variance formulas, enabling the implementation of PPI for a variety of estimands. 
Related directions include inverse-probability weighting (IPW)-style corrections for informative labeling/selection that interact with missingness \citep{carlson2025unifyingframeworkrobustefficient}, and imputation-powered inference (IPI) tailored to blockwise missingness that obtains valid estimation under MCAR and extensions (e.g., first-moment MCAR) \citep{ZhaoCandès2025IPI}. These works connect PPI-style ideas to classic missing-data semiparametric literature (e.g., \citealp{Robins1994,Rotnitzky1995,TsiatisMissingDataSemiparametricChapter}), which we will elaborate on later. Notably, the complexity of these semiparametric approaches may still pose challenges for practitioners without extensive statistical training, underscoring the need for practical tools and accessible workflows that bridge theory and application.

Taken together, these developments show that PPI is less a single estimator and more a general template: combine predictions and labeled outcomes via a bias-corrected objective, and then adapt that recipe to the structure of the problem at hand (e.g., efficiency constraints, selection mechanisms, and missingness patterns). Table~\ref{tab:ppi_summary} summarizes some representative variants of PPI, their core ideas, and improvements over the original PPI assumptions. Building on this discussion, we next situate PPI within the broader statistical literature, emphasizing its conceptual and methodological connections to existing approaches.

\section{Theoretical Connections with Statistics Literature}\label{sec:ppi_connection}

PPI connects naturally to several themes in modern statistics, including semiparametric theory, DR estimation, and data integration. A common thread is combining predictions from auxiliary models with limited gold-standard observations to achieve valid, efficient inference.

\subsection{Semiparametric efficiency and doubly robust estimation}

Prediction-powered inference can be understood through the lens of semiparametric efficiency theory as a particular augmented estimation strategy that leverages predictions to reduce variance when only a subset of outcomes are observed \citep{Robins1994, robins1995semiparametric, Rotnitzky1995}. Its robustness and efficiency properties depend critically on assumptions about the labeling mechanism.

Let the target parameter be a population risk
\(
\psi(\theta) = \mathbb{E}\{\ell(Y,X;\theta)\},
\)
where $\ell(\cdot)$ is a loss function and $\theta$ is the parameter of interest. Suppose outcomes are observed only when $S=1$, with labeling missing completely at random, so that
$S \perp (X,Y)$ and $\pi = P(S=1)$ is constant.

Under MCAR, the efficient influence function for $\psi(\theta)$ in the nonparametric model is
\[
\phi(\mathcal{O};\theta)
=
\frac{S}{\pi}\big\{\ell(Y,X;\theta) - \mu_\theta(X)\big\}
+ \mu_\theta(X) - \psi(\theta),
\]
where $\mathcal{O} = (X, S, SY)$ is the observed data and $\mu_\theta(X) = \mathbb{E}\{\ell(Y,X;\theta)\mid X\}$ is an arbitrary working model for the conditional expectation of the loss. Solving the corresponding estimating equation yields the familiar augmented estimator
\[
\widehat{\psi}_{\mathrm{AUG}}(\theta)
=
\frac{1}{n}\sum_{i=1}^n \mu_\theta(X_i)
+
\frac{1}{n}\sum_{i=1}^n \frac{S_i}{\pi}
\big\{\ell(Y_i,X_i;\theta) - \mu_\theta(X_i)\big\}.
\]
When $\mu_\theta(X)$ equals the true conditional expectation, this estimator achieves the semiparametric efficiency bound; for arbitrary $\mu_\theta(X)$, it remains unbiased under MCAR \citep{Bickel1993, vanderVaart1998}.

In more general missing-at-random  settings, where $P(S=1\mid X)=\pi(X)$ may vary with covariates, augmented estimators of this form are doubly robust: they remain consistent if either the outcome regression $\mu_\theta(X)$ or the labeling model $\pi(X)$ is correctly specified. This property, however, relies on explicitly modeling or knowing $\pi(X)$.

Standard PPI operates under the stronger MCAR assumption and sets $\pi = n_\ell / n$, a constant estimated labeling probability. As a result, the original PPI is not doubly robust in the usual MAR sense: if labeling depends on covariates, then consistency might be compromised. Rather, PPI should be viewed as an efficient augmented estimator, corresponding to a special case of the general influence-function-based framework.
PPI corresponds to the particular choices:
\(
\mu_\theta(X) = \ell(\hat Y(X), X; \theta),
\)
i.e., the loss evaluated at a predicted outcome and $\pi(X) = n_\ell/n$. This results in
\[
\widehat{\psi}_{\mathrm{AUG}}(\theta)
=
\frac{1}{n}\sum_{i=1}^n \ell(\hat Y_i, X_i; \theta)
-
\frac{1}{n_\ell}\sum_{i:S_i=1}
\big\{\ell(\hat Y_i, X_i; \theta) - \ell(Y_i, X_i; \theta)\big\},
\]
which coincides with the objective minimized by the PPI estimator. This representation makes explicit that PPI uses predictions as a variance-reducing augmentation rather than as a substitute for outcome data.

While the original PPI formulation does not explicitly exploit the full influence-function characterization, this perspective naturally motivates refinements. These include optimal augmentation, adjustment for non-constant labeling probabilities, and cross-fitting to mitigate overfitting-induced bias. Recent extensions incorporate these ideas to recover genuine doubly robust guarantees under MAR and to improve efficiency without narrowing the applicability of prediction-powered inference
\citep{ji2025predictions, carlson2025unifyingframeworkrobustefficient, testa2025semisup}. {Relatedly, \citet{xu2025unified} developed a unified semiparametric framework for semi-supervised inference that characterizes efficiency bounds and incorporates black-box predictions, thereby providing a broader theoretical perspective on PPI-style methods.}

\subsection{Control variates, survey sampling approaches, and finite-sample limits}

For mean estimation tasks, PPI can equivalently be framed as a variance reduction approach using control variates. In particular, for $\lambda \in \mathbb{R}$, letting $$U_{\lambda}= \lambda \cdot \Big( \frac{1}{n_u}\sum_{i:S_i=0} \hat Y_i -\frac{1}{n_\ell}\sum_{i : S_i = 1} \hat Y_i, \Big),$$ the PPI++ estimator can be rewritten as $\hat{\theta}_{\mathrm{PPI++}}(\lambda) = \hat{\theta}_{\mathrm{CC}} + U_{\lambda}$. Note that $U_{\lambda}$ is a quantity that has mean zero under Assumption (A1) and can thus be viewed as a control variate. Moreover, adding $U_{\lambda}$ to $\hat{\theta}_{\mathrm{CC}}$ preserves unbiasedness while reducing variance if $U_{\lambda}$ has sufficient negative correlation with $\hat{\theta}_{\mathrm{CC}}$. PPI++ optimizes the weight $\lambda \in [0,1]$ on this control-variate type adjustment to guarantee improvement over CC.


For estimation tasks beyond mean estimation, variants of PPI that involve debiasing estimators directly rather than empirical losses (e.g., \cite{gronsbell2024another,Kluger25GeneralizingPPI}) use the same point estimator originating in the literature on double-sampling (or two-phase sampling) designs \citep{ChenAndChen2000}. The estimators in these papers can also be (approximately) constructed by a control variate approach. In particular, these estimators can be written as $\hat{\theta}_{\mathrm{CC}} + U'$ where $U'$ is an approximate control variate in the sense that $\mathbb{E}[U']$ converges to $0$ asymptotically under Assumption (A1).

From a survey sampling perspective, PPI-based estimators for population means closely resemble classical model-assisted estimators, which use predictions from auxiliary models to improve efficiency while maintaining design-consistent inference. In particular, the original PPI estimator corresponds to the ``difference estimator'', and PPI++ aligns with the ``Generalized Regression (GREG) estimator'' when ML predictions $\hat{Y}$ are treated as auxiliary covariates \citep{sarndal2003model}. More generally, PPI parallels calibration estimation and pseudo-weighting methods used to combine non-probability and probability samples, particularly in the context of adjusting for sample bias using auxiliary information \citep{deville1992calibration, elliot2009combining, tsung2018model,KimEtAl2022}.

A key distinction, however, is that the survey sampling literature primarily focuses on methods for estimating population means or totals, whereas most papers in the PPI literature focus on general M-estimation problems, including estimating regression coefficients and other parameters defined through estimating equations. In addition, 
PPI is explicitly motivated by modern settings in which auxiliary information takes the form of black-box machine learning predictions, rather than low-dimensional, interpretable covariates typically used in classical survey approaches {(e.g., \cite{deville1992calibration,sarndal2003model}). We note that \citet{KimEtAl2022} also use machine learning predictions to encode the auxiliary information; however, they rely on different assumptions than PPI does. Their approach does not require labeled samples from the target population and instead requires that the machine learning predictions are sufficiently well-calibrated to transfer from the source to target populations. In contrast, for valid inference, PPI requires access to labels on a representative subsample of the target population but makes no assumptions about the quality of the black-box machine learning predictions. We point readers to \citet{mozer2026ppidifferenceestimatorrecognizing} for further discussion on the connections between the PPI literature and the survey sampling literature.
}


Finally, \citet{mani2025no} establish finite-sample limits: for Gaussian data, PPI++ outperforms CC only when $\left|\frac{\text{Cov}(Y,\hat{f}(X))}{\sqrt{\Var(Y)\Var(\hat{f}(X))}}\right| > 1/\sqrt{n_\ell - 2}$. Predictions must exceed a quality threshold depending on $n_\ell$; with small labeled samples, only highly accurate predictions help, and random predictions can {hurt} performance.

\subsection{Bayesian inference}\label{stat_connection:bayesian}
In Bayesian inference, several approaches closely parallel the logic of PPI by incorporating model-based predictions to enhance inference while maintaining validity. The catalytic prior \citep{huang2020catalytic, huang2025catalytic, li2024regularization} constructs a prior distribution by augmenting observed data with synthetic samples drawn from the predictive distribution of a simpler model trained on the same data. This effectively leverages model-based information to stabilize or regularize inference, much like PPI uses predictions to reduce variance without assuming model correctness. 


Similarly, Bayesian predictive inference emphasizes inference for population-level quantities rather than inference restricted to parameters of the observed sample. In this framework, a model is learned from the sampled data and used to generate predictions for non-sampled units, with uncertainty coherently quantified through the Bayesian posterior predictive distribution \citep{chen2010bayesian}. This perspective underlies recent work on Bayesian integration of probability and nonprobability samples, where auxiliary information from nonprobability surveys is incorporated through model-based predictions or priors to improve inference for population-level regression targets \citep{Salvatore2024bayesian}. The predictive paradigm has also been developed in survey methodology and finite-population inference, where predictions for unobserved units are treated as primary inferential targets rather than ancillary byproducts \citep{royall1970finite, valliant2000finite,williams2024improving}.

Together, these approaches illustrate how synthetic or predicted quantities can be systematically integrated into the Bayesian framework to enrich inference, mirroring the core idea of PPI in a probabilistic setting.

\subsection{Data integration and transfer learning}
Several recent frameworks in data integration and transfer learning share this fundamental philosophy. The synthetic data integration framework develops valid post-inference procedures by integrating external summary-level information, such as pre-trained models that can generate predictions given covariates, from external populations, by generating synthetic observations that reflect these external summaries \citep{gu2019synthetic, gu2023synthetic}. In the COMMUTE framework \citep{gu2023commute}, communication-efficient transfer learning is achieved across multiple sites by transforming local model information into synthetic data that can be used to calibrate predictions at a target site. Similarly, the Synthetic Volume Expansion framework explores how generative models can produce realistic synthetic datasets to improve downstream analysis, especially when privacy or access constraints limit raw data sharing \citep{shen2024boosting}.

These ideas echo classical approaches discussed earlier, such as GREG and calibration estimators in survey sampling, but are now generalized to accommodate modern challenges like black-box prediction models, high-dimensional features, and decentralized data.

\medskip
\noindent In summary, PPI is grounded in well-established statistical principles that clarify both its strengths and limitations. These theoretical foundations inform when PPI can be expected to improve upon classical approaches and highlight the assumptions that practitioners must carefully evaluate. We now turn from theory to practice, providing a structured workflow for implementing PPI in applied settings.

\section{Operationalizing PPI}\label{sec:guideline}

Sections~\ref{sec:ppi-framework} and \ref{sec:ppi_connection} clarify {why} PPI works and {when} its guarantees hold. However, translating these insights into practice requires systematic guidance: how should a practitioner determine whether PPI is appropriate for their study, which PPI variant should they choose, and how can they assess whether the core assumptions are plausible? To address these questions, we outline a three-step workflow for the valid application of PPI. This framework guides practitioners from initial feasibility assessment through implementation, helping identify when PPI is likely to yield efficiency gains and suggesting robust alternatives when key assumptions are partially violated.

\subsection{Step 1: Define the study and prepare data}

First, the identification of the research objective is essential, such as estimating a population mean, comparing treatment groups, fitting a regression model, or pursuing another inferential goal. The internal target dataset should then be constructed to include:

\begin{itemize}
\item a labeled subset of $n_\ell$ observations with gold-standard outcomes $(X_i, Y_i)$, and
\item an unlabeled subset of $n_u$ observations containing features only $(X_j)$.
\end{itemize}

A prediction model $\hat{f}$ for the generation of $\hat{Y}=\hat{f}(X)$ for all observations is also required. The development and training context of this model, including the data sources, should be carefully documented to facilitate assessment of the underlying assumptions in Step~2.

\subsection{Step 2: Assess assumptions and data quality}

The original PPI relies on assumptions (A1)--(A3) for validity, among which (A1) and (A2) are largely {untestable} in practice. Violations of these assumptions, such as MNAR mechanisms or {external training and internal inference overlap} (i.e., double-dipping), can lead to biased inference with $\mathbb{E}[\hat{\theta}_{\mathrm{PPI}}] \neq \theta$ or anti-conservative confidence intervals. For example, if prediction errors are smaller on labeled data than on unlabeled data due to distributional differences or overlap in { external labeled training and internal unlabeled inference samples}, the bias correction term will under-correct and propagate model bias into the final estimate. Consequently, practitioners should rely on study design and external knowledge, rather than statistical tests, to justify these assumptions.

Nonetheless, before applying PPI, it remains essential to systematically evaluate assumptions (A1)--(A3) using a combination of tools: (i) quantitative diagnostics on internal data, (ii) domain knowledge about the scientific context, and (iii) consultation with collaborators who understand data collection processes. Table~\ref{tab:assumption_diagnostics} summarizes recommended diagnostic strategies for assessing assumption plausibility in practice.

\paragraph{Assessing (A1):} 

An assessment should be conducted to determine whether the labeled and unlabeled samples plausibly arise from the same underlying distribution, as required under MCAR. Quantitative diagnostics such as standardized mean differences, Kolmogorov–Smirnov tests, or energy distance can be used to compare covariate distributions. Because outcomes are unobserved for the unlabeled sample, such tests must rely solely on $X$. When substantial differences are observed, it may indicate labeling bias or sampling frame shift, and practitioners should turn to variants designed for MAR or distributional shift (see Table~\ref{tab:ppi_summary} and Figure~\ref{fig:recommendation}). 

\paragraph{Assessing (A2):} 
It is important to verify that the prediction model was trained on data with no overlap with either the internal labeled or unlabeled samples. Any such overlap violates the required independence assumption and might lead to overly narrow confidence intervals and degraded coverage (see the discussion of “double-dipping” in Section~\ref{section:double_dipping}). Importantly, PPI may still be applied when the internal sample contains units drawn from the model’s original test set, provided that this test set was genuinely held out and not used for model training, tuning, or model selection.
When no external pre-trained model exists, or independence cannot be verified, use cross-fitting approaches such as Cross-PPI or Cross-PPBoot (see Table~\ref{tab:ppi_summary} and Figure~\ref{fig:recommendation}), or alternatively, semiparametric approaches for missing data can also be used directly \citep{TsiatisMissingDataSemiparametricChapter,Robins1994}.

\paragraph{Assessing (A3):} 
The extent and pattern of covariate missingness should be carefully quantified. Consistency of variable measurement, including units and coding schemes, should also be assessed across the labeled and unlabeled samples. When covariates are incomplete, imputation-powered inference or Predict-then-debias approaches may be considered (see Table~\ref{tab:ppi_summary} and Figure~\ref{fig:recommendation}).

\begin{table}[!htbp]
\centering
\caption{Diagnostic strategies for evaluating Prediction-Powered Inference (PPI) assumptions.}
\label{tab:assumption_diagnostics}
\begin{tabular}{@{}p{1.2cm}p{5.8cm}p{5.8cm}@{}}
\toprule
 & \textbf{Quantitative diagnostics} & \textbf{Domain/contextual evaluation} \\
\midrule
\textbf{(A1)} &
Compare covariate distributions {between labeled and unlabeled samples} using standardized mean differences, Kolmogorov–Smirnov tests, or energy distance. Substantial differences suggest violations of MCAR and may prompt the use of MAR-robust variants. &
Review data collection for selection biases; consult experts on whether differences reflect true heterogeneity or sampling artifacts. \\[0.6em]
\textbf{(A2)} &
Audit for sample overlap between { external labeled and internal unlabeled data}; examine systematic prediction shifts relative to observed outcomes in the labeled subset as a compatibility check. &
Confirm model provenance and external data sources; verify alignment of outcome definitions between { external data and internal target populations.} \\[0.6em]
\textbf{(A3)} &
Quantify missingness rates and patterns; perform sensitivity analyses under different imputation strategies. &
Verify measurement consistency; confirm variable definitions with data custodians. \\
\bottomrule
\end{tabular}
\end{table}

\subsection{Step 3: Select, implement, validate, and report PPI analyses}

\begin{figure}[!htbp]
    \centering
    \fbox{\includegraphics[width=.75\linewidth]{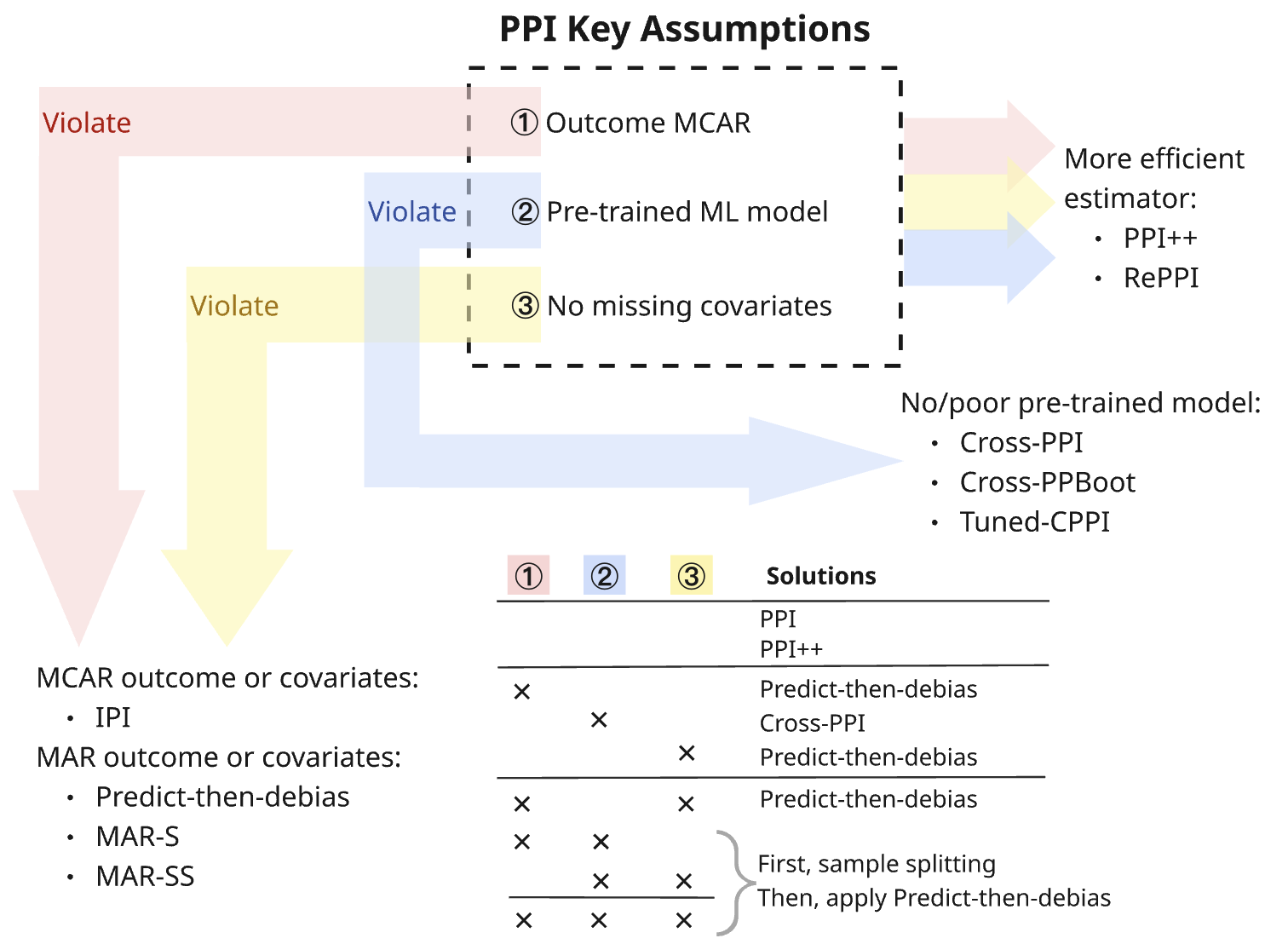}}
    \caption{Decision flowchart for selecting PPI variants based on assumption diagnostics.}
    \label{fig:recommendation}
\end{figure}

Based on findings from Step~2, select the PPI variant that addresses identified assumption violations. {The decision flowchart in Figure 2 is intended to organize existing PPI methods according to their assumptions and methodological innovations rather than prescribe a unique estimator for every inferential problem. Most PPI variants have been developed for broad classes of M-estimation problems, including means, regression coefficients, and generalized estimating equations, but some methods may currently provide theoretical guarantees or software implementations only for particular estimands or loss functions.} 
Figure~\ref{fig:recommendation} provides a decision flowchart:

\begin{itemize}[leftmargin=*]
    \item \textbf{All assumptions met:} Standard PPI or, for guaranteed efficiency gains, PPI++ or Re-PPI.
    \item \textbf{(A1) violated:} MAR-robust variants with propensity modeling or DR corrections.
    \item \textbf{(A2) violated:} e.g. Cross-PPI, Cross-PPBoot, or Tuned-CPPI with $K$-fold cross-fitting.
    \item \textbf{(A3) violated:} e.g. Imputation-powered inference or Predict-then-debias approaches.
    \item \textbf{Multiple violations:} Combined strategies such as cross-fitting with recalibration.
\end{itemize}

{\textit{Implementation considerations.} Several PPI variants, including PPI++, Re-PPI, and Tuned-CPPI, introduce data-adaptive tuning parameters or optimization procedures to improve efficiency. These quantities should generally not be selected manually. Instead, they are estimated automatically using the procedures proposed in the original papers, such as minimizing an estimated asymptotic variance or solving an auxiliary optimization problem, often with sample splitting or cross-fitting. Practitioners are therefore encouraged to use publicly available software implementations whenever possible rather than specifying tuning parameters heuristically. Likewise, when cross-fitting is employed (e.g., Cross-PPI and Cross-PPBoot), all stages of prediction model development, including preprocessing, feature selection, and hyperparameter tuning, should be confined to the training folds to avoid information leakage and preserve valid inference.}

After implementation, results should be validated by comparison with complete-case analyses to assess potential efficiency gains, and sensitivity analyses should be conducted to evaluate robustness. Transparent reporting is essential and should include clear documentation of how each assumption was assessed, which PPI variant was selected and for what reasons, and any limitations arising from assumptions that cannot be empirically verified.

{\textit{Reporting and interpretation.} Beyond selecting an appropriate PPI variant, transparent communication is essential for readers to evaluate the credibility of PPI-based inference. We recommend that applied analyses report four components.

First, investigators should clearly describe the prediction model, including how it was developed (external pre-trained model or cross-fitting), the source of the training data, and any procedures used to prevent overlap between model training and inference. Because independence between prediction and inference data cannot be verified statistically, documenting model provenance is critical.

Second, authors should explicitly justify the assumptions underlying the chosen PPI method. This includes describing why the labeled and unlabeled samples are expected to represent the same target population, summarizing diagnostic assessments of covariate comparability and missingness patterns, and explaining why the selected PPI variant is appropriate for the observed data structure.

Third, PPI results should be presented alongside an appropriate reference analysis, such as complete-case inference whenever feasible. Agreement between point estimates helps reassure readers that efficiency gains are not being achieved at the expense of changing the estimand, while differences in confidence interval width illustrate the practical value of incorporating predictions.

Finally, efficiency gains themselves should be interpreted cautiously. Narrower confidence intervals indicate that the prediction model contributes useful auxiliary information, but they should not be viewed as evidence that the prediction model is accurate or scientifically validated. Conversely, little or no efficiency improvement does not invalidate PPI; rather, it suggests that the available predictions contribute limited additional information beyond the labeled data. In this sense, PPI separates predictive performance from inferential validity: prediction quality affects efficiency, whereas validity depends primarily on whether the underlying assumptions are satisfied.}

\section{Empirical Evaluation with MOSAIKS Housing Price}\label{sec:DataBasedExperiments}

The MOSAIKS housing dataset \citep{rolf2021a, rolf2021b, proctor2023parameter} contains $N=46{,}418$ observations with housing price, income, nightlights, and road length. For all four variables, predictions from a satellite-imagery model are available for all units, while the true values are also observed for the full dataset. This structure allows us to impose outcome missingness by design, treat a subset of observations as labeled, and use the remaining units as unlabeled. The availability of the full ground truth further enables direct comparison between PPI-based estimators and oracle estimators computed from the complete data. In this section, the target parameter is the vector of linear regression coefficients relating housing price ($Y$) to income, nightlights, and road length ($X$).

\subsection{Comparing PPI-based estimators for linear regression coefficients}

\begin{figure}[!htbp] 
    \centering
    \begin{subfigure}[t]{0.49\textwidth}
        \centering
        \includegraphics[height=1.6in]{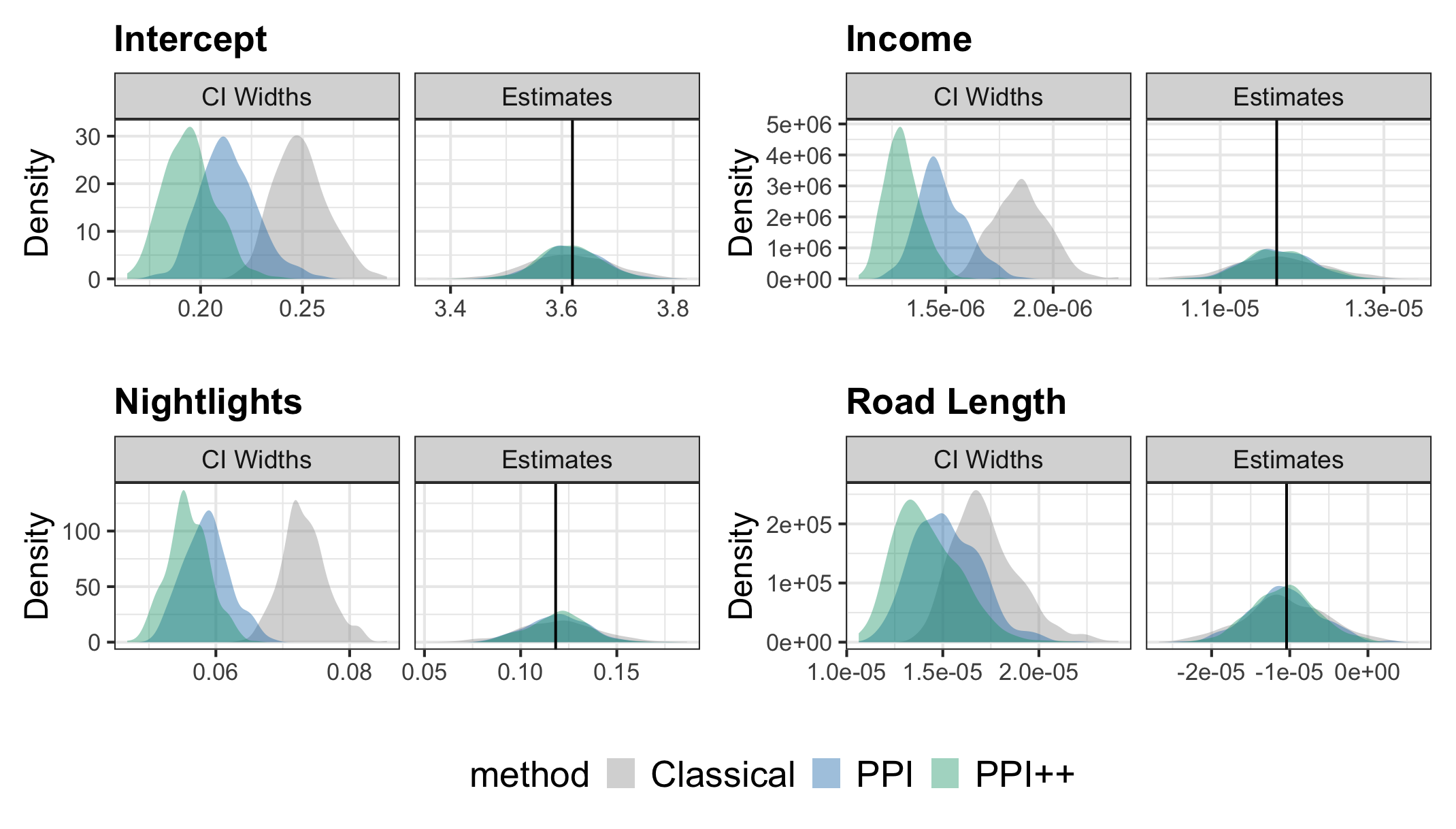}
        \caption{Methods: PPI, PPI++, and classical.}
    \end{subfigure}\hfill
    \begin{subfigure}[t]{0.49\textwidth}
        \centering
        \includegraphics[height=1.6in]{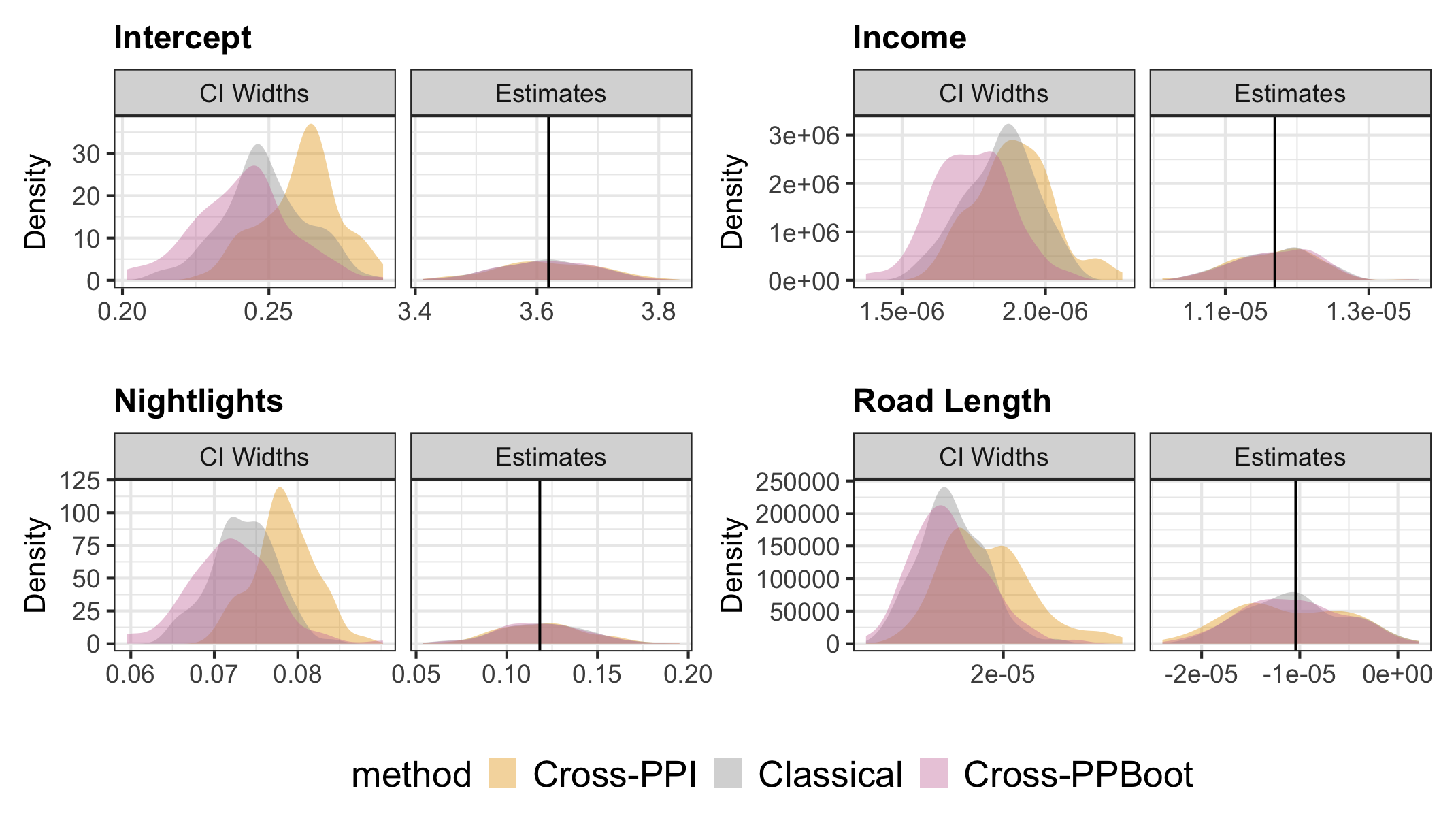}
        \caption{Methods: Cross-PPI, Cross-PPBoot and classical. XGBoost prediction model with income, nightlights, and road length}
    \end{subfigure}

    \vspace{0.5em}
    
    \begin{subfigure}[t]{0.49\textwidth}
        \centering
        \includegraphics[height=1.6in]{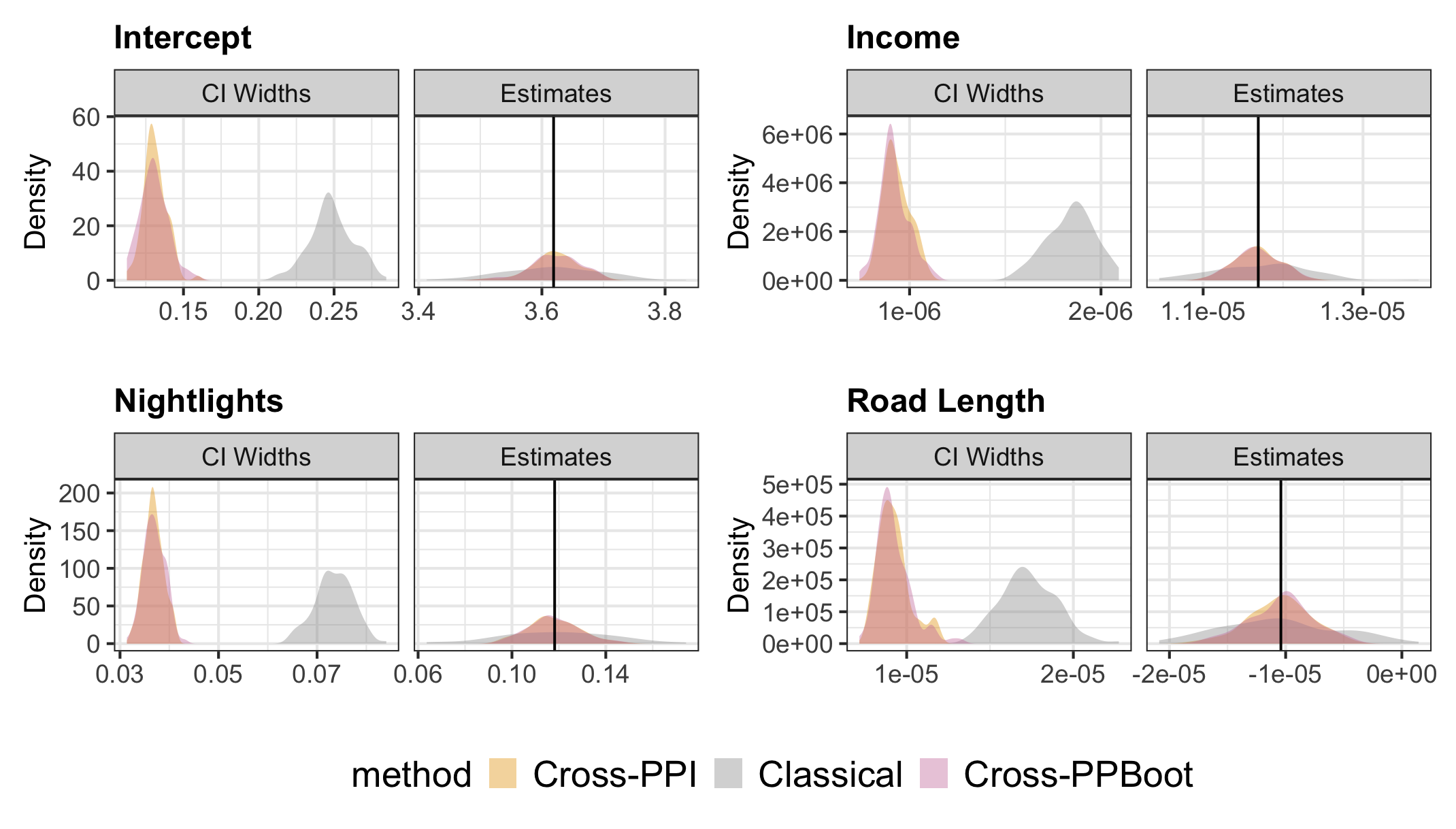}
        \caption{Methods: Cross-PPI, Cross-PPBoot, and classical. XGBoost prediction model with income, nightlights, road length, and an additional covariate highly correlated with housing price.}
    \end{subfigure}\hfill
    \begin{subfigure}[t]{0.49\textwidth}
        \centering
        \includegraphics[height=1.6in]{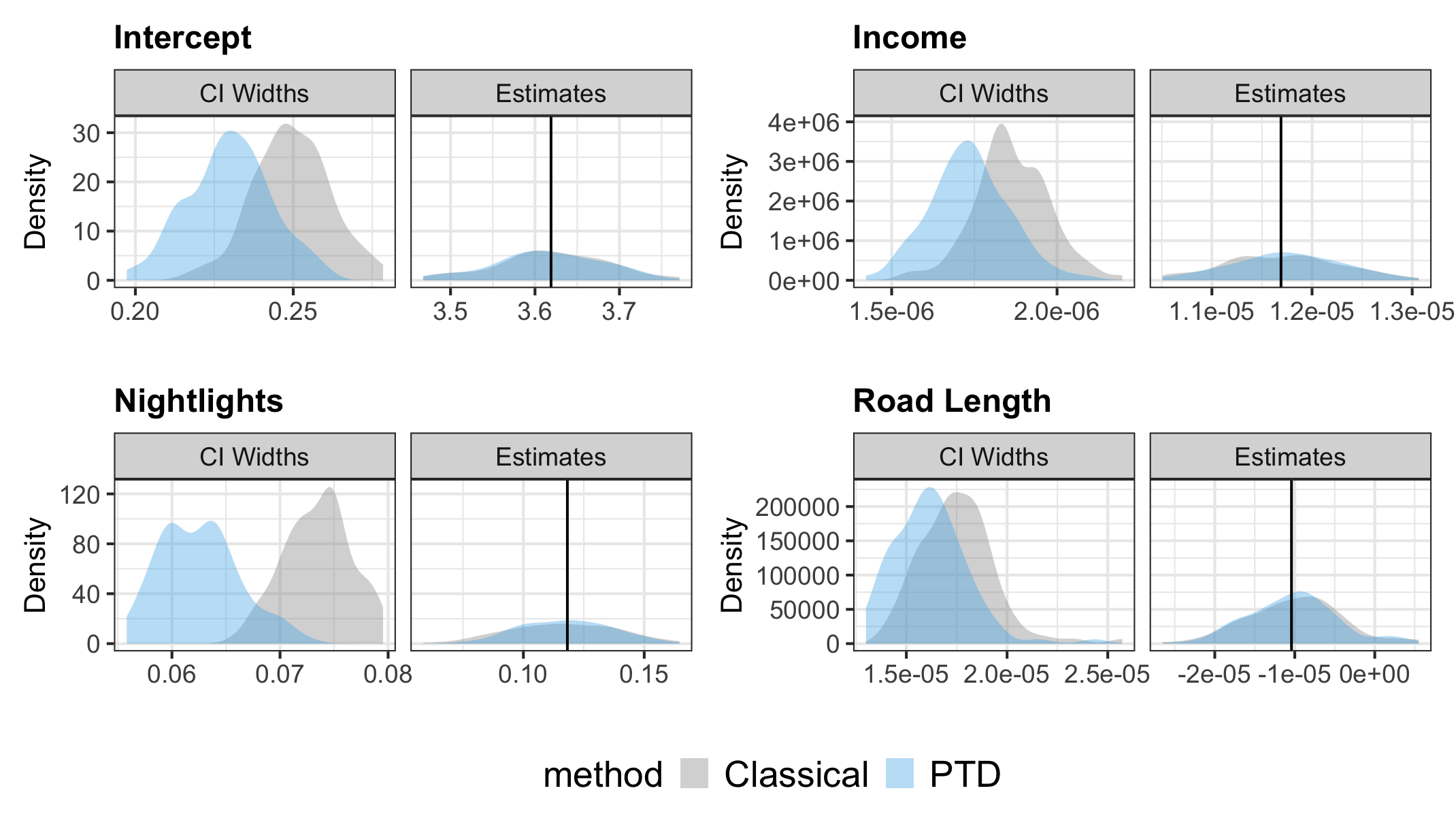}
        \caption{Methods: Predict-then-debias and classical.}
    \end{subfigure}%
    
    \caption{Estimates and average 90\% confidence interval length for linear regression coefficients in MOSAIKS housing data. The vertical line indicates the true coefficient values. }\label{fig:Mosaiks}
\end{figure}

We first evaluate multiple PPI-based estimators using the MOSAIKS housing dataset. From the full dataset of $N=46,418$ observations, we repeatedly sample 1,000 labeled units and treat the remaining observations as unlabeled, performing 500 Monte Carlo replications. For each method, we summarize the mean point estimates and the average 90\% confidence interval lengths across replications to assess bias and efficiency (\cref{fig:Mosaiks}). \cref{fig:Mosaiks}(a) compares the performance of the classical approach with labeled data only, PPI, and PPI++. For Cross-PPI, in the absence of an external pre-trained model, we use a five-fold cross-fitted XGBoost model based on income, nightlights, and road length to obtain out-of-fold predictions (\cref{fig:Mosaiks}(b)). To examine how prediction accuracy influences the performance of Cross-PPI and Cross-PPBoot, we simulate an additional covariate strongly correlated with the outcome and repeat the analysis using this expanded feature set in the XGBoost model for prediction (\cref{fig:Mosaiks}(c)). For the Predict-then-debias approach, we rely on the pre-trained predictions $\hat{X}$ (i.e., pre-trained predictions of income, nightlights and road length) and $\hat{Y}$ (i.e., pre-trained predictions of housing price) provided in the dataset (\cref{fig:Mosaiks}(d)).


Across 500 replications, the standard PPI and PPI++ procedures consistently produced narrower confidence intervals than classical analysis while retaining essentially unbiased point estimates, confirming their ability to extract efficiency gains even when prediction quality is modest. In contrast, the Cross-PPI and Cross-PPBoot methods did not uniformly improve efficiency relative to the classical approach: their interval widths were comparable to, or sometimes larger than, the classical intervals when the predictor was limited to the three original covariates (income, nightlights, and road length). We suspect that future methods, which introduce a power tuning parameter (as used in PPI++) into the Cross-PPI and Cross-PPBoot frameworks, could prevent such scenarios where the classical intervals are narrower. Recent work on Tuned-CPPI \citep{sifaou2024semi} represents an important step in this direction, demonstrating the potential benefits of adaptively selecting such tuning parameters. Notably, when we introduced an additional covariate that was strongly predictive of the outcome and incorporated it into the cross-fitted XGBoost model, both Cross-PPI and Cross-PPBoot exhibited substantial confidence interval shrinkage. Introducing the strongly predictive covariate also led to improvements over PPI++ from the earlier setting in \cref{fig:Mosaiks}(a), where less accurate predictions of housing price were used, demonstrating that these methods can yield meaningful efficiency gains when high-quality predictions are available. The Predict-then-debias estimator showed stable performance throughout, offering modest confidence interval reductions relative to classical inference. We suspect that these gains would have been more substantial had some subset of the covariates (income, nightlights, and road length) been fully available for all 46,418 observations in the dataset.

\subsection{The effect of ``double-dipping'' on inference validity} \label{section:double_dipping}

``Double-dipping'' refers to the use of overlapping data for both training the prediction model and performing inference, i.e. when the same units appear in both the training set used to fit the prediction model and the labeled subset used to estimate parameters via PPI. This overlap can induce bias, lead to overfitting, and result in understated uncertainty quantification. 

To investigate the impact of double-dipping, we focus on PPI++ applied to the MOSAIKS housing data. We train an XGBoost model to predict housing price from income, nightlights, road length, and an additional simulated covariate. Assume practitioners have access to an external labeled data $\mathcal{D}^{\text{ex}}$ and an internal data $\mathcal{D}^{\text{in}}$ with both labeled and unlabeled observations. We consider two scenarios: (1) In the {double-dipping} setting, the model is trained on the combined labeled data from both $\mathcal{D}^{\text{in}}$ and $\mathcal{D}^{\text{ex}}$ and predictions are generated for all data in both sets; and (2) In the {holdout} setting, the prediction model is only trained on $\mathcal{D}^{\text{ex}}$ and predictions are obtained in $\mathcal{D}^{\text{in}}$. We then estimate the linear regression coefficients for housing price using $\mathcal{D}^{\text{in}}$, comparing the performance of PPI and PPI++ under the double-dipping and holdout regimes. We repeat this process for 200 simulation runs. For each combination of labeled sample sizes $n_\ell^{\text{ex}}$ and $n_\ell^{\text{in}}$ from $\mathcal{D}^{\text{ex}}$ and $\mathcal{D}^{\text{in}}$, respectively, we evaluate the empirical coverage and width of the 90\% confidence interval. Note that here\footnote{ In the notation of Section \ref{sec:ppi-framework}, $n^{\text{in}}_\ell$ and $n^{\text{in}}_u$ correspond to $n_l$ and $n_u$, and thus $n^{\text{in}}=n^{in}_\ell+n^{in}_u=n$.} we have $n^{\text{ex}}=n^{\text{ex}}_\ell$ since all external data are labeled, $n^{\text{in}}=N-n^{\text{ex}}$, and the sample size of unlabeled data in the internal data $n^{\text{in}}_u=n^{\text{in}}-n^{\text{in}}_\ell$. 

\begin{table}[h]
\centering 

\caption{Empirical coverage and average width of the 90\% confidence interval obtained by the classical, PPI, and PPI++ approaches with double-dipping or holdout setting under different labeled sample sizes.} \label{tab:doubledipping}

\resizebox{\textwidth}{!}{
\begin{tabular}{|ccc|cc|cc|cc|}
\hline
\multicolumn{3}{|c|}{($n_\ell^{\text{ex}}, n_\ell^{\text{in}}$)}                                                                                                 & \multicolumn{2}{c|}{(23000,1000)} & \multicolumn{2}{c|}{(4600,1000)} & \multicolumn{2}{c|}{(1000,1000)} \\ \hline
\multicolumn{1}{|c|}{Coef}                          & \multicolumn{1}{c|}{Source}                          & Method    & \multicolumn{1}{c|}{Coverage}                      & Width                           & \multicolumn{1}{c|}{Coverage}                     & Width                           & \multicolumn{1}{c|}{Coverage}                     & Width                           \\ \hline
\multicolumn{1}{|c|}{\multirow{5}{*}{Intercept}}    & \multicolumn{1}{c|}{}                                & Classical & \multicolumn{1}{c|}{0.910}                          & 0.249                           & \multicolumn{1}{c|}{0.900}                          & 0.249                           & \multicolumn{1}{c|}{0.875}                        & 0.250                            \\ \cline{2-9} 
\multicolumn{1}{|c|}{}                              & \multicolumn{1}{c|}{\multirow{2}{*}{Double Dipping}} & PPI       & \multicolumn{1}{c|}{0.915}                         & 0.192                           & \multicolumn{1}{c|}{0.725}                        & 0.138                           & \multicolumn{1}{c|}{0.470}                         & 0.0845                         \\ \cline{3-9} 
\multicolumn{1}{|c|}{}                              & \multicolumn{1}{c|}{}                                & PPI++     & \multicolumn{1}{c|}{0.915}                         & 0.192                           & \multicolumn{1}{c|}{0.725}                        & 0.138                           & \multicolumn{1}{c|}{0.470}                         & 0.0845                          \\ \cline{2-9} 
\multicolumn{1}{|c|}{}                              & \multicolumn{1}{c|}{\multirow{2}{*}{Holdout}}        & PPI       & \multicolumn{1}{c|}{0.920}                          & 0.234                           & \multicolumn{1}{c|}{0.895}                        & 0.248                           & \multicolumn{1}{c|}{0.900}                          & 0.257                           \\ \cline{3-9} 
\multicolumn{1}{|c|}{}                              & \multicolumn{1}{c|}{}                                & PPI++     & \multicolumn{1}{c|}{0.925}                         & 0.231                           & \multicolumn{1}{c|}{0.895}                        & 0.235                           & \multicolumn{1}{c|}{0.880}                         & 0.236                           \\ \hline
\multicolumn{1}{|c|}{\multirow{5}{*}{Income ($\times 10^{6}$)}}     
& \multicolumn{1}{c|}{}                                & Classical & \multicolumn{1}{c|}{0.910}                          & $1.840$                    & \multicolumn{1}{c|}{0.910}                         & $1.820$                     & \multicolumn{1}{c|}{0.890}                         & $1.850$                      \\ \cline{2-9} 
\multicolumn{1}{|c|}{}                              & \multicolumn{1}{c|}{\multirow{2}{*}{Double Dipping}} & PPI       & \multicolumn{1}{c|}{0.915}                         & $1.380$                     & \multicolumn{1}{c|}{0.830}                         & $0.973$                     & \multicolumn{1}{c|}{0.510}                         & $0.586$                     \\ \cline{3-9} 
\multicolumn{1}{|c|}{}                              & \multicolumn{1}{c|}{}                                & PPI++     & \multicolumn{1}{c|}{0.915}                         & $1.380$                       & \multicolumn{1}{c|}{0.830}                         & $0.974$                     & \multicolumn{1}{c|}{0.510}                         & $0.586$                    \\ \cline{2-9} 
\multicolumn{1}{|c|}{}                              & \multicolumn{1}{c|}{\multirow{2}{*}{Holdout}}        & PPI       & \multicolumn{1}{c|}{0.930}                          & $1.690$                      & \multicolumn{1}{c|}{0.915}                        & $1.790$                      & \multicolumn{1}{c|}{0.940}                         & $1.860$                       \\ \cline{3-9} 
\multicolumn{1}{|c|}{}                              & \multicolumn{1}{c|}{}                                & PPI++     & \multicolumn{1}{c|}{0.935}                         & $1.670$                       & \multicolumn{1}{c|}{0.900}                          & $1.680$                       & \multicolumn{1}{c|}{0.890}                         & $1.700$                        \\ \hline
\multicolumn{1}{|c|}{\multirow{5}{*}{Nightlights}} & \multicolumn{1}{c|}{}                                & Classical & \multicolumn{1}{c|}{0.925}                         & 0.0733                          & \multicolumn{1}{c|}{0.895}                        & 0.0735                          & \multicolumn{1}{c|}{0.880}                         & 0.0733                          \\ \cline{2-9} 
\multicolumn{1}{|c|}{}                              & \multicolumn{1}{c|}{\multirow{2}{*}{Double Dipping}} & PPI       & \multicolumn{1}{c|}{0.920}                          & 0.0567                          & \multicolumn{1}{c|}{0.700}                          & 0.0403                          & \multicolumn{1}{c|}{0.540}                         & 0.0241                          \\ \cline{3-9} 
\multicolumn{1}{|c|}{}                              & \multicolumn{1}{c|}{}                                & PPI++     & \multicolumn{1}{c|}{0.920}                          & 0.0568                          & \multicolumn{1}{c|}{0.700}                          & 0.0403                          & \multicolumn{1}{c|}{0.540}                         & 0.0241                          \\ \cline{2-9} 
\multicolumn{1}{|c|}{}                              & \multicolumn{1}{c|}{\multirow{2}{*}{Holdout}}        & PPI       & \multicolumn{1}{c|}{0.925}                         & 0.0701                          & \multicolumn{1}{c|}{0.905}                        & 0.0742                           & \multicolumn{1}{c|}{0.890}                         & 0.0774                          \\ \cline{3-9} 
\multicolumn{1}{|c|}{}                              & \multicolumn{1}{c|}{}                                & PPI++     & \multicolumn{1}{c|}{0.915}                         & 0.0693                          & \multicolumn{1}{c|}{0.895}                        & 0.0709                          & \multicolumn{1}{c|}{0.885}                        & 0.0713                          \\ \hline
\multicolumn{1}{|c|}{\multirow{5}{*}{Road Length ($\times 10^{5}$)}}  & \multicolumn{1}{c|}{}                                & Classical & \multicolumn{1}{c|}{0.905}                         & $1.740$                        & \multicolumn{1}{c|}{0.940}                         & 1.740                       & \multicolumn{1}{c|}{0.945}                        & 1.720                       \\ \cline{2-9} 
\multicolumn{1}{|c|}{}                              & \multicolumn{1}{c|}{\multirow{2}{*}{Double Dipping}} & PPI       & \multicolumn{1}{c|}{0.930}                          & 1.340                       & \multicolumn{1}{c|}{0.925}                        & 0.911                      & \multicolumn{1}{c|}{0.880}                         & 0.547                      \\ \cline{3-9} 
\multicolumn{1}{|c|}{}                              & \multicolumn{1}{c|}{}                                & PPI++     & \multicolumn{1}{c|}{0.930}                          & 1.340                       & \multicolumn{1}{c|}{0.925}                        & 0.912                      & \multicolumn{1}{c|}{0.880}                         & 0.547                      \\ \cline{2-9} 
\multicolumn{1}{|c|}{}                              & \multicolumn{1}{c|}{\multirow{2}{*}{Holdout}}        & PPI       & \multicolumn{1}{c|}{0.950}                          & 1.700                        & \multicolumn{1}{c|}{0.935}                        & 1.810                       & \multicolumn{1}{c|}{0.965}                        & 1.910                       \\ \cline{3-9} 
\multicolumn{1}{|c|}{}                              & \multicolumn{1}{c|}{}                                & PPI++     & \multicolumn{1}{c|}{0.945}                         & 1.680                       & \multicolumn{1}{c|}{0.915}                        & 1.720                       & \multicolumn{1}{c|}{0.950}                         & 1.710                       \\ \hline
\end{tabular}
}

\end{table}

Table~\ref{tab:doubledipping} shows that the double-dipping strategy
produces confidence intervals that are systematically too narrow across all
regression coefficients. Under double-dipping, coverage for both PPI and
PPI++ deteriorates sharply as the external labeled training sample size decreases. For example,
coverage for the coefficients of income and nightlights drops
to roughly 50\% when $(n_\ell^{\text{ex}}, n_\ell^{\text{in}}) = (1000, 1000)$, accompanied by
interval widths that are less than half of those produced by the classical
approach. This reflects substantial overconfidence caused by reusing the same
observations for both prediction and inference. 

In contrast, the holdout strategy yields intervals with coverage values near the nominal 90\% level across all sample sizes, and PPI++ yields a more efficient confidence interval than the classical intervals. Overall, these results emphasize that PPI and PPI++ require sample splitting or cross-fitting, if no pre-trained model is available, to avoid optimistic inference, as reusing predictions from the same data substantially understates variability.

\subsection{Failure of PPI-based methods under the missing not at random setting}

When outcomes are MNAR, the key identification assumptions required by all PPI-based methods are violated. To study this setting, we construct an MNAR labeling mechanism in which the probability of being labeled depends directly on the outcome itself: units with housing price, $Y$, above a specified percentile cutoff are made more likely to be labeled than units below the cutoff. {We vary both the cutoff and the strength of this selection, considering the $60$th- and $80$th-percentile thresholds and labeling weight multipliers of $2\times$, $5\times$, and $10\times$ for observations above the threshold. Concretely, observations above the cutoff are assigned sampling weights $2$, $5$, or $10$, whereas observations below the cutoff are assigned weight one, and the labeled sample is then drawn without replacement with probabilities proportional to these weights. This induces dependence between the labeling probability and the outcome, violating the MCAR and MAR assumptions and creating systematic differences between the labeled and unlabeled samples.} Under this setup, we evaluate the performance of the classical approach, PPI, PPI++, Cross-PPI, Cross-PPBoot, and Predict-then-debias for estimating linear regression coefficients.

\begin{figure}[htp] 
    \centering
    \includegraphics[width=\linewidth]{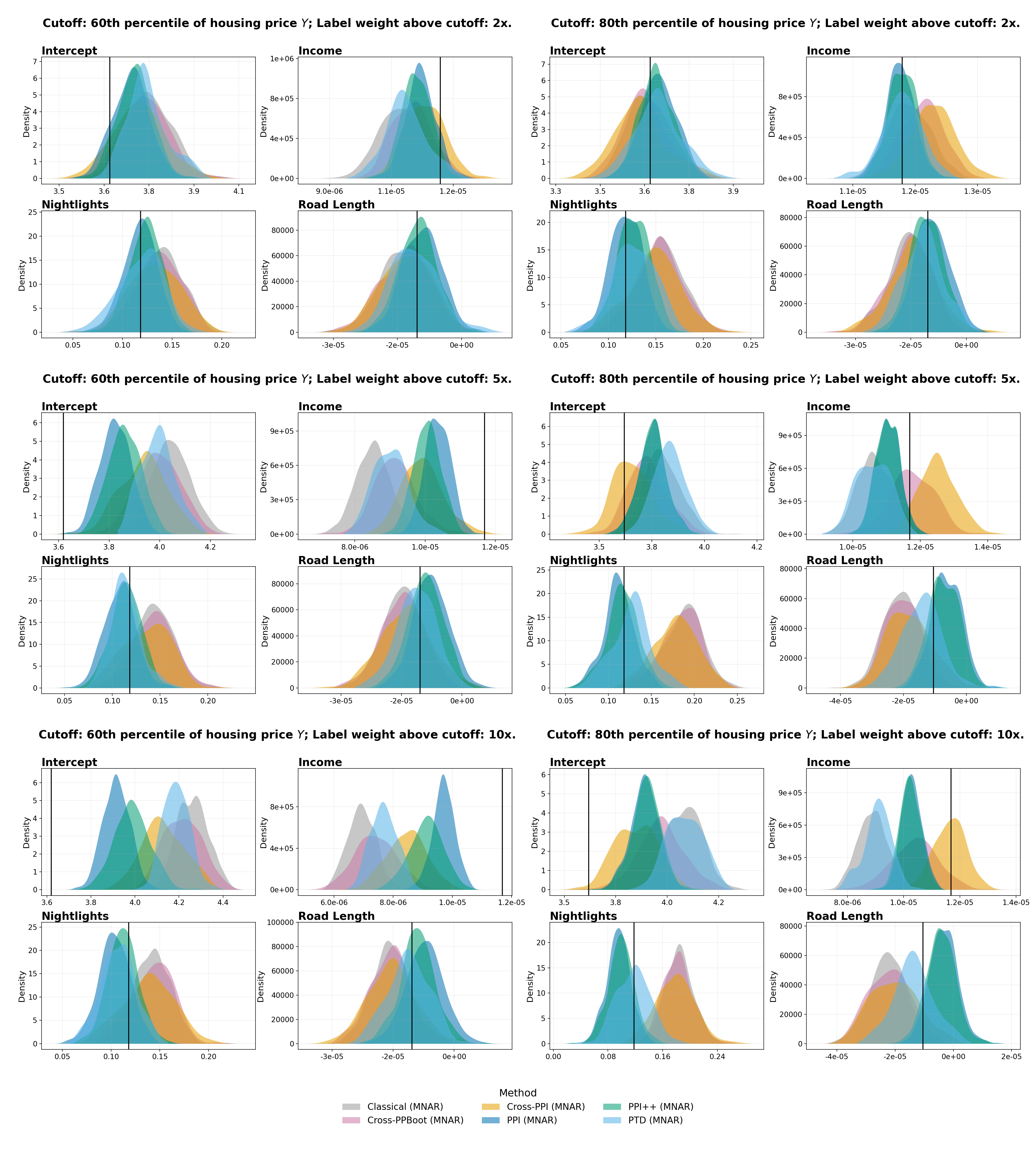}
    \caption{MNAR assessment under outcome-dependent labeling. In each setting, observations with housing price $Y$ above the stated percentile threshold are made more likely to be labeled by the multiplier shown in the subtitle. Columns compare the 60th- and 80th-percentile cutoffs, and rows compare labeling weight multipliers ($2\times$, $5\times$, and $10\times$). Within each setting, the four coefficient panels correspond to the Intercept, Income, Nightlights, and Road Length terms, and the shaded curves show the sampling distributions of the corresponding estimates across simulation replicates for the Classical, Cross-PPI, Cross-PPBoot, PPI, PPI++, and PTD methods. The vertical black line marks the true value of each coefficient.}\label{fig:MNAR}
\end{figure}


Under this setup, none of the estimators remain valid. As shown in Figure~\ref{fig:MNAR}, all methods produce biased estimates that deviate from the truth, although the magnitude and pattern of bias vary across coefficients and across MNAR settings. {For a fixed percentile cutoff, the bias generally becomes more pronounced as the labeling weight multiplier increases, since stronger outcome-dependent selection induces a more severe departure between the labeled and unlabeled samples. By contrast, within the range considered here, changing the percentile cutoff does not produce a clear systematic pattern in bias across methods and coefficients.} The classical approach is also biased, since it relies on a labeled subset whose selection depends directly on the outcome. Occasional alignment with the truth is incidental rather than systematic. These results confirm that outcome-driven missingness invalidates both classical and PPI-type estimators, and underscore the need for methods that explicitly account for MNAR mechanisms.

\section{Discussion \& Conclusion}\label{sec:discussion}

In this paper, we demystify PPI by presenting its core intuition,  formal assumptions, and recent methodological advances in Section~\ref{sec:ppi-framework}, and highlighting its connections to established statistical theory in Section~\ref{sec:ppi_connection}. We provided a practical three-step workflow (Section~\ref{sec:guideline}) and a consolidated summary of representative methods (Table~\ref{tab:ppi_summary}) to support the applied use. Additionally, we illustrated the performances of various PPI methods using the MOSAIKS housing data in Section \ref{sec:DataBasedExperiments}. 

Since its introduction, PPI has evolved from a single method into a diverse family of procedures spanning efficiency-guaranteed variants, MAR-aware generalizations, cross-fitting extensions, and federated implementations. We conclude by organizing our assessment around a strengths, weaknesses, opportunities, and threats (SWOT) framework. \emph{Strengths} highlight what makes PPI valuable in its current form, while \emph{Weaknesses} identify limitations that can plausibly be overcome through methodological advancement or improved practice. \emph{Opportunities} point to promising directions for future development and application, while \emph{Threats} denote more fundamental issues---such as violations of untestable assumptions---that can compromise validity in ways that may be difficult or impossible to fully resolve.

\subsection{Strengths}

PPI's central strength lies in its principled use of predictions to extract information from unlabeled data while preserving valid inference: predictions are exploited but never treated as ground truth. Unlike post-prediction inference frameworks that implicitly require correct model specification \citep{motwani2024revisiting}, PPI maintains validity even when the prediction model is misspecified---a crucial property when leveraging complex ML models whose error distributions are difficult to characterize.

When unlabeled data are plentiful and predictions are reasonably informative, PPI delivers tighter confidence intervals than complete-case analysis. PPI++ strengthens this with formal asymptotic efficiency guarantees \citep{PPI++}, ensuring that incorporating predictions never increases asymptotic variance relative to ignoring them entirely. The framework's modularity has enabled ready extension across diverse settings: cross-fitting for internal model training \citep{CrossPPIPaper_ZrnicCandes}, stratification for heterogeneous prediction quality \citep{StratifiedPPI}, federated variants for distributed data governance \citep{luo2024federated}, and MAR-robust versions for non-random labeling \citep{testa2025semisup, ying2025covshift,carlson2025unifyingframeworkrobustefficient}.

PPI also benefits from deep connections to established statistical theory. As discussed in Section~\ref{sec:ppi_connection}, the framework can be understood through the lens of semiparametric efficiency, doubly robust estimation, control variates, and model-assisted survey sampling, etc. These connections provide both theoretical grounding and practical guidance: insights from decades of missing-data research directly inform when and how PPI can be expected to work well.

Finally, the growing body of successful applications---spanning biomedicine \citep{mccaw2024synthetic}, remote sensing \citep{lu2025remotesensing}, social science \citep{BroskaEtAl}, and natural language processing (NLP) evaluation \citep{StratifiedPPI}---demonstrates that PPI's theoretical promise translates into real-world impact.

\subsection{Weaknesses}

Several limitations of current PPI methodology can plausibly be addressed through continued research and improved practice.

\textit{No universal efficiency guarantee in the original formulation.}
The original PPI estimator does not guarantee efficiency gains over complete-case analysis; improvements depend on prediction quality and can reverse when predictions are poor. While PPI++ resolves this asymptotically through adaptive tuning \citep{PPI++}, finite-sample results show that predictions must exceed a quality threshold that depends on the labeled sample size \citep{mani2025no}. Clearer practical guidance on assessing whether predictions are ``good enough''---beyond the Gaussian-specific bounds currently available---would help practitioners make informed decisions.

\textit{Complexity and the usability gap.}
The proliferation of PPI variants---each with distinct assumptions, tuning procedures, and implementation requirements---poses a barrier to adoption. Practitioners face a daunting landscape: PPI, PPI++, Cross-PPI, Cross-PPBoot, stratified variants, MAR-robust extensions, and more (Table~\ref{tab:ppi_summary}). Without accessible software, unified interfaces, and clear documentation, researchers may select inappropriate methods or misinterpret outputs. The decision flowchart (Figure~\ref{fig:recommendation}) and practical workflow (Section~\ref{sec:guideline}) provided here represent initial steps, but comprehensive tooling with built-in diagnostics remains an important gap.

\textit{Limited guidance on assumption diagnostics.}
While we outline diagnostic strategies in Table~\ref{tab:assumption_diagnostics}, current guidance remains somewhat ad hoc. Standardized procedures for assessing covariate balance, detecting distribution shift, and evaluating prediction calibration across subgroups---ideally integrated into PPI software---would help practitioners identify potential assumption violations before they compromise inference.

\textit{Computational considerations underexplored.}
The computational requirements of different PPI variants are rarely discussed. For large-scale applications with millions of unlabeled observations, understanding the computational trade-offs between methods would help practitioners make feasible choices.

\subsection{Opportunities}

Several directions promise continued methodological and practical growth.

\textit{Federated and privacy-preserving inference.}
Federated PPI enables valid inference when labeled data cannot be centralized \citep{luo2024federated}, addressing privacy constraints increasingly common in healthcare, finance, and multi-institutional research. As data governance requirements tighten globally, methods that provide PPI's benefits without requiring data pooling will become increasingly valuable.

\textit{Active inference and adaptive labeling.}
Active inference \citep{ZrnicActiveInference} and other adaptive labeling schemes \citep{kluger2026mestimationtwophasemultiwavesampling} connect PPI to experimental design by strategically acquiring labels that result in reduced estimator variance. This offers clear benefits in resource-limited settings where labeling is expensive: rather than labeling observations uniformly at random, practitioners can target observations where labels provide maximal information. {Stability and efficiency improvements of adaptive labeling strategies within and beyond these frameworks as well as integration of semi-parametrically efficient PPI variants with adaptive labeling approaches represents a promising frontier.} 

{\textit{Multiple predictions of the missing variables.} The PPI methods discussed in this paper focus on settings where there is a single function $\hat{f}$ that is used to predict the missing labels. In settings where there are instead multiple predictions of each missing label (perhaps from different machine learning models), further efficiency gains are possible. Recent works develop methods that leverage multiple predictions in such settings and demonstrate efficiency gains compared to PPI methods that use a single prediction model \citep{cowen-breen2026multiplepredictionpowered,shan2025sada,chen2025surrogatepoweredinferenceregularizationadaptivity,byun2025valid}.}


\textit{Unifying semiparametric frameworks.}
Recent work has begun unifying PPI with classical semiparametric theory under MAR and decaying-overlap regimes \citep{testa2025semisup, ying2025covshift, chen2025unified}, delivering doubly robust estimators that maintain validity under partial model misspecification. Reinterpreting predictions as surrogates further clarifies conditions for optimal efficiency within the PPI class \citep{ji2025predictions}. Continued theoretical development will sharpen understanding of when and how much PPI can help.

\textit{Domain-specific methodological advances.}
Tailoring PPI to domain-specific challenges has yielded important advances: remote-sensing applications in environmental science \citep{lu2025remotesensing}, synthetic surrogates for genome-wide association studies \citep{mccaw2024synthetic,miao2024vgwas}, and hybrid human-LLM evaluation in NLP \citep{StratifiedPPI, BroskaEtAl}. Each domain brings unique data structures, missingness patterns, and prediction models; continued collaboration between methodologists and domain experts will expand PPI's reach.

\textit{Software ecosystem and educational resources.}
As the ecosystem of variants grows, developing centralized software with unified interfaces, benchmarking pipelines for method comparison, and educational materials for diverse audiences will be essential to lower barriers to adoption and promote responsible use.

\subsection{Threats}

Beyond limitations that can be overcome through methodological improvement, several more fundamental issues can compromise PPI's validity in ways that warrant serious caution.

\textit{Fundamentally untestable assumptions.}
The core assumptions underlying PPI---particularly (A1) and (A2)---cannot be fully verified from observed data. The MCAR assumption concerns the relationship between labeling and \emph{unobserved} outcomes, which is inherently unverifiable: we can compare covariate distributions between labeled and unlabeled samples, but this provides no direct evidence about whether labeling depends on $Y$ given $X$. Similarly, confirming that {external labeled training and internal unlabeled inference data} are truly independent requires external documentation rather than statistical diagnostics. This creates a risk of false confidence: practitioners may conduct diagnostic checks, obtain reassuring results, and incorrectly conclude that all assumptions hold. We emphasize that such diagnostics are necessary but not sufficient---justification for assumptions must ultimately rest on domain knowledge and study design.

\textit{Opaque model provenance.}
The increasing use of foundation models and large language models whose training corpora are undocumented or proprietary poses a fundamental threat to assumption (A2). When practitioners cannot verify what data were used to train the prediction model, the independence assumption becomes effectively uncheckable. If the prediction model was trained on data that overlaps with the inference sample---even partially or indirectly---validity guarantees may be silently compromised. This threat is particularly acute in domains like genomics and NLP where large shared databases are common and foundation models are trained on internet-scale corpora. Unlike the ``double-dipping'' scenarios in Section~\ref{sec:DataBasedExperiments} where contamination was known by construction, real-world contamination may be undetectable.

\textit{Unobserved drift and MNAR mechanisms.}
While covariate diagnostics can reveal some forms of distribution shift between labeled and unlabeled samples, shifts in the conditional distribution $P(Y \mid X)$ are fundamentally undetectable without observing $Y$ in both populations. If the relationship between outcomes and covariates differs systematically between labeled and unlabeled observations---due to temporal drift, geographic variation, or selection on unobservables---the bias correction estimated on labeled data will not properly correct predictions on unlabeled data. The most severe case arises under MNAR mechanisms, where labeling depends directly on $Y$ even after conditioning on $X$. As demonstrated in Section~\ref{sec:DataBasedExperiments}, all PPI-based methods fail under MNAR, as do classical complete-case approaches. MAR-robust variants offer protection when labeling depends only on observables, but they cannot address shifts in the outcome model itself, nor can any amount of methodological sophistication within the PPI framework rescue inference when labeling is outcome-dependent. Sensitivity analysis approaches for MNAR, well-developed in the missing data literature, have not yet been systematically integrated with PPI.

\textit{Positivity violations and sparse overlap.}
{The bias correction in MAR-robust PPI variants} relies on labeled observations adequately representing the regions of covariate space where predictions are applied. When the positivity (overlap) condition is violated---meaning certain subpopulations have few or no labeled observations---bias correction becomes unreliable precisely where it may be most needed. This threat manifests in two ways. First, predictions may be systematically miscalibrated for underrepresented subgroups, yet the labeled sample provides insufficient information to detect or correct this miscalibration. Second, asymptotic guarantees may provide false comfort: even when assumptions hold in principle, finite-sample bias correction can be inadequate if the labeled sample poorly covers regions where predictions are inaccurate. Subgroup analyses and calibration diagnostics can help identify such problems, but they cannot fully protect against them, particularly when the relevant subgroups are defined by unobserved or high-dimensional characteristics.

{\textit{Transparent reporting and interpretation.} Clear reporting standards are essential for the responsible use of PPI. Applied analyses should document the prediction model and its provenance, justify the assumptions underlying the selected PPI variant, summarize diagnostic assessments, and compare PPI results with conventional complete-case analyses whenever feasible. Efficiency gains should be interpreted as reflecting the additional information contributed by the predictions, rather than as evidence that the prediction model is scientifically validated. Such reporting will improve the transparency, reproducibility, and interpretability of PPI-based analyses.}

\subsection{Concluding Remarks}

PPI exemplifies a broader movement in modern statistics toward integrating ML predictions into principled inferential frameworks. While recent development has been driven largely by the ML community, the foundational ideas build on long-standing practices in survey sampling (model-assisted estimation, calibration, poststratification) and missing data analysis (augmented inverse probability weighting, doubly robust estimation) that similarly leverage auxiliary models to enhance inference. What distinguishes PPI is its explicit design for settings where a labeled subsample enables nonparametric bias correction without distributional assumptions about prediction errors.

The guiding principle remains simple: extract information from predictions, then guard against their errors. With careful attention to assumptions---justified through domain knowledge and study design rather than statistical tests alone---and judicious selection among the growing family of PPI variants, practitioners can achieve both improved precision and valid inference. We hope this synthesis, together with the workflow in Section~\ref{sec:guideline} and the summary in Table~\ref{tab:ppi_summary}, enables researchers across disciplines to responsibly harness the predictive power of modern ML for rigorous statistical analysis.

\newpage

\begin{table}[H]
\centering
\scriptsize
\renewcommand{\arraystretch}{1.5}   
\caption{\centering
Summary of representative PPI methods organized according to the four major development axes discussed in Section~\ref{sec:ppi-framework}. The improvement axes correspond to improvements in (i) efficiency guarantees (Efficiency), (ii) relaxing Assumption (A1): distribution comparability between labeled and unlabeled samples (typically extending from MCAR to MAR), (iii) relaxing Assumption (A2): eliminating the requirement for an independent externally pre-trained prediction model, and (iv) relaxing Assumption (A3): allowing incomplete covariate information or more general missingness patterns. $\checkmark$ indicates that the corresponding method extends the original PPI along that axis.
}
\label{tab:ppi_summary}
\resizebox{\textwidth}{!}{%
\begin{tabular}{
  >{\raggedright\arraybackslash}m{0.15\textwidth}
  >{\raggedright\arraybackslash}m{0.31\textwidth}
  c c c c
  >{\raggedright\arraybackslash}m{0.30\textwidth}
}
\toprule
 \multirow{2}{*}{\textbf{Method}}
&
\multirow{2}{=}{\centering\textbf{Core idea}}
&
\multicolumn{4}{c}{\textbf{Improvement Axes}}
&
\multirow{2}{=}{\centering\textbf{Remarks}}
\\
\cmidrule(lr){3-6}
&
&
\textcolor{efficiencyblue}{\textbf{Efficiency}}
&
\textcolor{aoneorange}{\textbf{A1}}
&
\textcolor{atwogreen}{\textbf{A2}}
&
\textcolor{athreepurple}{\textbf{A3}}
&
\\

\midrule

PPI
\newline\citep{PPI}
&
Use predictions on unlabeled cases, subtract their bias using labeled cases to obtain valid estimations of the target parameter without trusting the model
&
&
&
&
&
Baseline PPI estimator.
\\

PPI++
\newline\citep{PPI++}
&
Add a tuning parameter to the original PPI estimator to achieve guaranteed improvement (asymptotically) over classical intervals using only the labeled data
&
\AxisCheck{efficiencyblue}
&
&
&
&
Optimization of a scalar tuning parameter $\lambda$.
\\

Cross-PPI
\newline\citep{CrossPPIPaper_ZrnicCandes}
&
Cross-fitting style: impute via ML, then debias across folds to maintain validity
&
&
&
\AxisCheck{atwogreen}
&
&
Removes the need for an external pretrained model via cross-fitting.
\\

MAR-S
\newline\citep{carlson2025unifyingframeworkrobustefficient}
&
Semiparametric/DR procedures under MAR with known propensity scores; unifies text/image predictors into structured targets
&
\AxisCheck{efficiencyblue}
&
\AxisCheck{aoneorange}
&
&
\AxisCheck{athreepurple}\textsuperscript{*}
&
Known propensity scores; semiparametric and doubly robust.
\\

Chen et.al
\newline\citep{chen2025unified}
&
General $Z$-estimation with general missingness patterns under MAR using ML imputations; pattern-specific weighting
&
\AxisCheck{efficiencyblue}
&
\AxisCheck{aoneorange}
&
&
\AxisCheck{athreepurple}
&
Pattern-stratified $Z$-estimation for arbitrary missingness.
\\

Re-PPI
\newline\citep{ji2025predictions}
&
Re-calibrate an imputed loss linked to a surrogate/predicted outcome; optimize efficiency over PPI-class even with imperfect loss estimation via sample splitting.
&
\AxisCheck{efficiencyblue}
&
&
&
\AxisCheck{athreepurple}\textsuperscript{*}
&
Tuning involves function-valued optimization; optimal within the PPI class.
\\

MAR-SS
\newline\citep{testa2025semisup}
&
Allows the labeling mechanism to be estimated; retains double robustness under MAR, permits decaying overlap asymptotically.
&
\AxisCheck{efficiencyblue}
&
\AxisCheck{aoneorange}
&
\AxisCheck{atwogreen}
&
\AxisCheck{athreepurple}\textsuperscript{*}
&
Estimated propensity scores; doubly robust with decaying overlap.
\\

Cross-PPBoot
\newline\citep{Zrnic2024PPBoot}
&
Uses $K$-fold out-of-fold predictions and bootstrap to form CIs without relying on an external pre-trained model.
&
&
&
\AxisCheck{atwogreen}
&
&
Cross-fitting with bootstrap confidence intervals.
\\

Tuned-CPPI
\newline\citep{sifaou2024semi}
&
Refines Cross-PPI by tuning how unlabeled data influences correction while preserving cross-fitting independence.
&
\AxisCheck{efficiencyblue}
&
&
\AxisCheck{atwogreen}
&
&
Cross-fitting with adaptive tuning for improved efficiency.
\\

IPI
\newline\citep{ZhaoCandès2025IPI}
&
Use ML imputation to construct estimating equations that remain valid; covers MCAR and extensions (e.g., first-moment MCAR).
&
\AxisCheck{efficiencyblue}
&
\AxisCheck{aoneorange}
&
\AxisCheck{atwogreen}
&
\AxisCheck{athreepurple}
&
Imputation-powered estimating equations.
\\

Predict-then-debias
\newline\citep{Kluger25GeneralizingPPI}
&
Use ML-predicted covariates (on large unlabeled set) with a small complete sample to debias downstream inference; uses bootstrap to construct valid CIs
&
\AxisCheck{efficiencyblue}
&
\AxisCheck{aoneorange}
&
&
\AxisCheck{athreepurple}
&
Tuning involves matrix-valued optimization; handles imputed covariates and complex sampling.
\\

\bottomrule
\end{tabular}
}
\vspace{0.35em}
\parbox{\textwidth}{\tiny \textsuperscript{*} These papers do not empirically evaluate their methods under A3 violations; the checkmark with an asterisk indicates that the authors briefly discuss extensions beyond A3 or that their framework appears extendable in principle.}
\end{table}

\newpage
\bibliography{references}

\end{document}